%% file: main.tex
\documentclass[10pt]{article} 

\newcommand{\submissionstage}{preprint}

\usepackage{etoolbox}
\newif\ifblind
\ifdefstring{\submissionstage}{review}{\blindtrue}{\blindfalse}
\ifdefstring{\submissionstage}{accepted}{\usepackage[accepted]{tmlr}}{%
\ifdefstring{\submissionstage}{preprint}{\usepackage[preprint]{tmlr}}{%
\usepackage{tmlr}}}

\input{math_commands.tex}

\usepackage{keyval}

\usepackage{graphicx}
\usepackage{xcolor}
\usepackage{tikz}
\usetikzlibrary{
    calc, 
    arrows.meta,
    patterns,
    patterns.meta,
    positioning,
    matrix,
    shapes,
    fit
}
\usepackage{pgfplots}
\pgfplotsset{compat=1.18}
\usetikzlibrary{external}
\usetikzlibrary{arrows.meta, positioning, shapes.callouts}
\usepackage{pgfplotstable}
\usepackage{svg}

\usepackage{bfhcolors}
\input{./style/plot_style}
\input{./style/graph_style}

\usepackage{wrapfig}
\usepackage{caption}
\usepackage{subfig}

\usepackage{booktabs}

\usepackage{silence}

\usepackage{amssymb}

\usepackage{algorithm}
\usepackage{algpseudocode}
\usepackage{listings}

\usepackage{xspace}

\usepackage{stfloats}
\usepackage{placeins}
\usepackage{paracol}
\usepackage{multirow}

\usepackage{enumitem}

\usepackage[colorlinks=true,
            linkcolor=black,
            citecolor=black,
            urlcolor=black,
            filecolor=black,
            bookmarks=true]{hyperref}
\usepackage{url}
\hypersetup{linktocpage}

\usepackage{natbib}
\usepackage[acronym]{glossaries}
\usepackage[hyperpageref]{backref}
\renewcommand{\backref}[1]{}
\renewcommand{\backrefalt}[4]{%
  \ifcase #1
\hspace{-1mm}%
  \or
    [cited on page~#2]%
  \else
    [cited on pages~#2]%
  \fi}
\glsdisablehyper

\newcommand{\task}[1]{\texttt{#1}}
\newcommand{\op}[1]{\texttt{#1}}
\newcommand{\llm}[1]{\texttt{#1}}

\newcommand{\acrlonglower}[1]{%
    \expandafter\MakeLowercase\expandafter{\glsentrylong{#1}}%
}
\newcommand{\acrlonglowerpl}[1]{%
    \expandafter\MakeLowercase\expandafter{\glsentrylongpl{#1}}%
}

\newcommand{\LLM}{\acrshort{llm}\xspace}
\newcommand{\GoO}{\acrlonglower{goo}\xspace}
\newcommand{\GoOs}{\acrlonglowerpl{goo}\xspace}
\newcommand{\GoT}{\acrlonglower{got}\xspace}
\newcommand{\RL}{\acrlonglower{rl}\xspace}
\newcommand{\ML}{\acrlonglower{ml}\xspace}

\title{Reinforced Graph of Thoughts:\\RL-Driven Adaptive Prompting for LLMs}

\author{
        \name\hspace{-1.4mm}Manuel Noah Riesen \email manuelnoah.riesen@students.bfh.ch \\
        \addr School of Engineering and Computer Science \\
        Bern University of Applied Sciences \\
        Bern, Switzerland
        \AND
        \name Peter Alfred von Niederhäusern \email peter.vonniederhaeusern@bfh.ch \\
        \addr School of Engineering and Computer Science \\
        Bern University of Applied Sciences \\
        Bern, Switzerland
}

\def\month{05}  
\def\year{2026} 
\def\openreview{\url{https://openreview.net/forum?id=XXXX}} 

\begin{document}

\input{glossary}

\maketitle

\begin{abstract}
\input{./sections/0_abstract}
\end{abstract}

\clearpage

\input{./sections/1_introduction}

\input{./sections/2_background}

\clearpage

\input{./sections/3_methodology}

\input{./sections/4_results}

\input{./sections/5_discussion}

\clearpage





\phantomsection
\addcontentsline{toc}{section}{References}
\bibliography{references}
\bibliographystyle{tmlr}

\pdfbookmark[0]{Appendices}{appendices-bookmark}
\appendix
\include{./sections/6_appendix}

\end{document}

%% file: math_commands.tex
\usepackage{amsmath,amsfonts,bm}

\def\eqref#1{equation~\ref{#1}}

\def\1{\bm{1}}

\DeclareMathAlphabet{\mathsfit}{\encodingdefault}{\sfdefault}{m}{sl}
\SetMathAlphabet{\mathsfit}{bold}{\encodingdefault}{\sfdefault}{bx}{n}



%% file: style/plot_style.tex
\colorlet{primary}{green!80!black}
\colorlet{secondary}{blue!80!black}
\colorlet{accent}{magenta!80!black}
\colorlet{negative}{red!50}

\definecolor{category a}{RGB}{27,158,119}
\definecolor{category b}{RGB}{217,95,2}
\definecolor{category c}{RGB}{117,112,179}

\pgfplotscreateplotcyclelist{mark categories}{
    {color=category a, mark=square*},
    {color=category b, mark=triangle*},
    {color=category c, mark=o},
}

\pgfplotscreateplotcyclelist{negative}{
    {fill=negative}
}

\pgfplotscreateplotcyclelist{bar categories original}{
    {blue,fill=blue!30!white,mark=none},
    {red,fill=red!30!white,mark=none},
    {green,fill=green!30!white,mark=none},
    {orange,fill=orange!30!white,mark=none},
    {purple,fill=purple!30!white,mark=none},
}

\tikzdeclarepattern{
    name=lines a,
    parameters={
        \pgfkeysvalueof{/pgf/pattern keys/size},
        \pgfkeysvalueof{/pgf/pattern keys/angle},
        \pgfkeysvalueof{/pgf/pattern keys/line width},
    },
    bounding box={
        (0,-0.5*\pgfkeysvalueof{/pgf/pattern keys/line width}) and
        (\pgfkeysvalueof{/pgf/pattern keys/size}, 0.5*\pgfkeysvalueof{/pgf/pattern keys/line width})
    },
    tile size={(\pgfkeysvalueof{/pgf/pattern keys/size},
    \pgfkeysvalueof{/pgf/pattern keys/size})},
    tile transformation={rotate=\pgfkeysvalueof{/pgf/pattern keys/angle}},
    defaults={
        size/.initial=10pt,
        angle/.initial=45,
        line width/.initial=0.4pt,
    },
    code={
      \draw [line width=\pgfkeysvalueof{/pgf/pattern keys/line width}]
        (0,0) -- (\pgfkeysvalueof{/pgf/pattern keys/size},0);
    },
}

\pgfplotscreateplotcyclelist{bar categories}{
    {
        category a,
        fill=category a!30!white,
        mark=none,
        postaction={
            pattern=lines a,
            pattern color=category a
        }
    },
    {
        category b,
        fill=category b!30!white,
        mark=none,
        postaction={
            pattern=dots,
            pattern color=category b
        }
    },
    {
        category c,
        fill=category c!30!white,
        mark=none,
        postaction={
            pattern=crosshatch,
            pattern color=category c
        }
    }
}

\pgfplotscreateplotcyclelist{line categories}{
    {color=category a, line width=0.8pt},
    {color=category b, line width=0.8pt},
    {color=category c, line width=0.8pt}
}

\pgfplotscreateplotcyclelist{training metrics categories}{
    {category a},
    {category b},
    {category c},
}

%% file: style/graph_style.tex
\tikzset{
    positive thought/.style={
        ellipse,
        rounded corners,
        draw=gray,
        fill=BFH-LightGreen!25!white,
        thick,
        inner sep=5pt,
        minimum width=1cm,
        minimum height=0.75cm
    },
    negative thought/.style={
        ellipse,
        rounded corners,
        draw=gray,
        fill=BFH-LightRed!25!white,
        thick,
        inner sep=5pt,
        minimum width=1cm,
        minimum height=0.75cm
    },
    unscored thought/.style={
        ellipse,
        rounded corners,
        draw=gray,
        fill=BFH-LightBlue!25!white,
        thick,
        inner sep=5pt,
        minimum width=1cm,
        minimum height=0.75cm
    },
    operation/.style={
        draw=gray,
        fill=BFH-LightBlue!25!white,
        thick,
        inner sep=5pt
    },
    thought/.style={
        rectangle,
        rounded corners,
        draw=gray,
        dashed,
        fill=BFH-LightPurple!25!white,
        thick,
        inner sep=5pt
    },
}

%% file: glossary.tex
\newacronym{rgot}{RGoT}{Reinforced Graph of Thoughts}
\newacronym{ci}{CI}{Continuous Integration}
\newacronym{ml}{ML}{Machine Learning}
\newacronym{io}{IO}{Input-Output}
\newacronym{cot}{CoT}{Chain of Thought}
\newacronym{tot}{ToT}{Tree of Thoughts}
\newacronym[shortplural=GoTs,longplural=Graphs of Thoughts]{got}{GoT}{Graph of Thoughts}
\newacronym{kgot}{KGoT}{Knowledge Graph of Thoughts}
\newacronym[shortplural=GoOs,longplural=Graphs of Operations]{goo}{GoO}{Graph of Operations}
\newacronym{llm}{LLM}{Large Language Model}
\newacronym{rl}{RL}{Reinforcement Learning}
\newacronym{prng}{PRNG}{pseudo-random number generator}
\newacronym{api}{API}{Application Programming Interface}
\newacronym{json}{JSON}{JavaScript Object Notation}
\newacronym{dql}{DQL}{Deep Q Learning}
\newacronym{dqn}{DQN}{Deep-Q-Network}
\newacronym{a2c}{A2C}{Advantage Actor Critic}
\newacronym{ppo}{PPO}{Proximal Policy Optimization}
\newacronym{td}{TD}{Temporal-Difference}
\newacronym{mdp}{MDP}{Markov Decision Process}
\newacronym{rag}{RAG}{Retrieval Augmented Generation}
\newacronym{dsl}{DSL}{Domain-Specific Language}
\newacronym{cpi}{CPI}{conservative policy iteration}

\newglossaryentry{tf}
{
        name=training front,
        description={The front of the training complexities. Indicates that the succeeding complexities are unknown to an agent}
}
\newglossaryentry{pytorch}
{
        name=PyTorch,
        description={A Python library for Machine Learning and Tensor computation}
}
\newglossaryentry{gym}
{
        name=Gymnasium,
        description={API standard for Reinforcement Learning environment definition}
}
\newglossaryentry{sb3}
{
        name=Stable Baselines3,
        description={A collection of implementations of Reinforcement Learning algorithms in PyTorch}
}
\newglossaryentry{sb3_contrib}{
    name=Stable Baselines3 Contrib,
    description={Repository with experimental code for Stable Baselines3}
}
\newglossaryentry{pypi}{
    name=PyPI,
    description={Python Package Index}
}

%% file: sections/0_abstract.tex
Graph of Thoughts (GoT), a generalized form of recent prompting paradigms for large language models (LLMs), has been shown to be useful for elaborate problem solving.
By executing a graph of operations, thoughts of the LLM are structured as an arbitrary graph, forming the actual graph of thoughts.
Originally, the graph of operations is defined manually, which requires in-depth knowledge about the solution of the problem to solve.
Such a static graph of operations is rigid and therefore lacks adaptability.
We propose Reinforced Graph of Thoughts (RGoT), an automated approach to the GoT prompting paradigm that leverages reinforcement learning (RL) to adaptively generate a graph of operations from a human-defined set.
Results indicate that, under certain constraints, it is possible to construct graphs of operations adaptively to the task’s complexity in an automated way. 

\vspace{0.25cm}

\noindent
Code repository: \\ 
\ifblind
    <blinded for review>
\else
\url{https://github.com/mriesen/reinforced-graph-of-thoughts}
\fi

Data on request.

\vspace{0.25cm}

%% file: sections/1_introduction.tex
\section{Introduction}
The use of advanced prompting paradigms such as \acrfull{cot} and \acrfull{tot} has been shown to be useful when it comes to deliberate problem solving with \acrfullpl{llm} 
\citep{wei_chain_2022} \citep{yao_tree_2024}.

\acrfull{got} represents a generalized form of such prompting paradigms.
The thoughts of an \LLM are modeled as an arbitrary graph \citep{besta_graph_2024}.


We propose \acrfull{rgot} - an automated approach of \acrshort{got} to enhance both accessibility and adaptability of complex prompting paradigms.

\paragraph{Accessibility}
Leveraging more advanced prompting paradigms such as \acrshort{got} comes with the drawback of an increased complexity for the user.
Not only is it important to formulate the prompts correctly, but the operations must be composed in the right order as well.
By reducing the user’s work to only the definition of the operations, the \acrlong{got} paradigm becomes more accessible and can be adapted more easily.

\paragraph{Adaptability}
The \acrlong{got} framework by \citet{besta_graph_2024} involves the construction of a complete and static \GoO. Such a graph is rigid in its behavior.
Therefore, a static \GoO may only be applicable to input with strictly defined properties.
Moreover, a static graph is unable to adapt to a language model's nondeterministic output, which harms the system's resilience against the language model's entropic nature.
Removing the rigidity from the \GoO by letting an agent construct the graph continuously in an automated fashion, one may overcome such restrictions to a certain degree.

\paragraph{Differentiation to other Approaches}
While there is another work on automating \acrshort{got} \citep{ha_auto_2024}, we present a more sophisticated approach that does not rely on logical capabilities of the underlying language model to build a \GoO.
We evaluate on similar tasks as \citet{besta_graph_2024} and \citet{ha_auto_2024} to demonstrate coverage, not to directly compare results, as the approaches differ fundamentally in adaptability.

In contrast to approaches that use graphs to structure multi-agent collaboration or external knowledge retrieval - such as \acrfull{kgot} \citep{besta_kgot_2025} - our method targets the automated construction of the reasoning topology itself.

%% file: sections/2_background.tex
\section{Background}

\subsection{Prompting Paradigms}

\begin{figure*}[ht]
    \centering
    \includegraphics[width=0.65\textwidth]{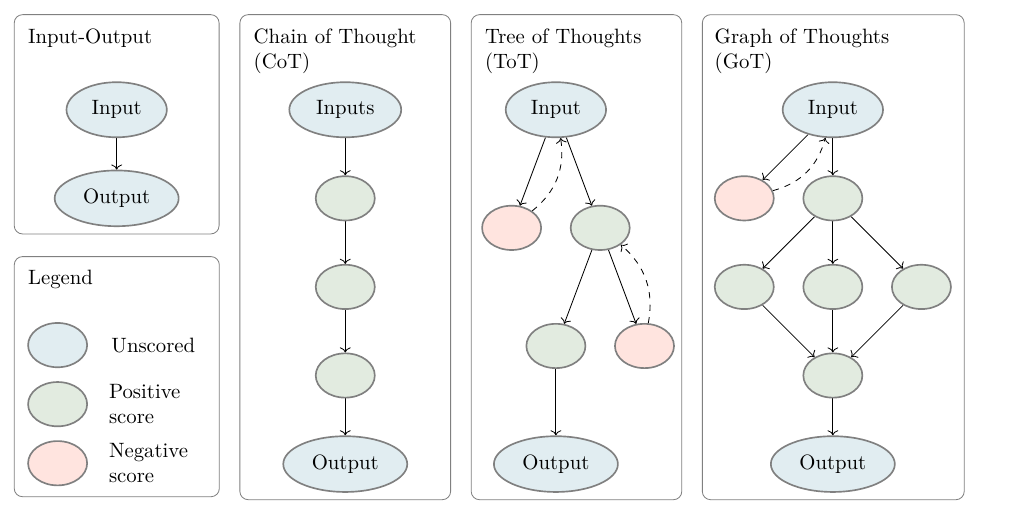}
    \caption[Prompting Paradigms]{Prompting Paradigms with increasing complexity (from left to right).}
    \label{fig:prompting_paradigms}
\end{figure*}

\paragraph{\texorpdfstring{\acrlong{io}}{Input-Output} Prompting}
The most basic prompting paradigm is the \acrfull{io} prompting.
It describes the interaction between a client (user or system) and a language model, where the client sends a single prompt that is then answered by the language model by a single result (thought).

\paragraph{\texorpdfstring{\acrlong{cot}}{Chain-of-Thought}}
\acrfull{cot} \citep{wei_chain_2022} introduces intermediate reasoning steps in the process of the interaction between the client and the language model.
It has been shown that prompting language models to write down the intermediate steps to solve a problem can be beneficial. 
Besides the benefit of the transparency in the reasoning process, language models may perform better when asked to solve a task step by step.

\paragraph{\texorpdfstring{\acrlong{tot}}{Tree of Thoughts}}
\acrfull{tot} \citep{yao_tree_2024} introduces a more sophisticated structure for the resulting thoughts of a language model.
With \acrshort{tot}, the thoughts are structured as a tree.
This allows looking ahead and backtracking within the tree, so that one can navigate to the desired result.

\paragraph{\texorpdfstring{\acrlong{got}}{Graph of Thoughts}}
\acrfull{got} \citep{besta_graph_2024} represents a generalized form of prompting paradigms such as \acrshort{cot} and \acrshort{tot}.
Thoughts of the language model are structured as an arbitrary graph.
To be precise, \acrshort{got} is introduced as a framework with its own \acrshort{api} definition.
A graph of operations acts as the controlling structure. In the original definition, the \GoO is filled with thoughts until the desired result is achieved.
A key novelty with a graph structure in comparison to a tree is the ability to aggregate thoughts.
This is highly beneficial for tasks that can be split into simpler subtasks. The subtasks can then be solved locally. Afterwards, the results can be aggregated back into a single combined result. 
It is also possible to represent preceding prompting structures as graphs of thoughts.

\subsection{\texorpdfstring{\acrlong{rl}}{Reinforcement Learning}}
\acrfull{rl} is a subfield of \ML in which an agent learns by the interaction with an environment 
\citep{sutton_reinforcement_2018}.
The problem can be expressed as a \acrfull{mdp}.
Formally, a \acrshort{mdp} is represented by a tuple \((\mathcal{S}, \mathcal{A}, \mathcal{P}, \mathcal{R}, \gamma)\) where \(\mathcal{S}\) is the environment's state space, \(\mathcal{A}\) is the agent's action space, \(\mathcal{P}\) is a transition probability function, \(\mathcal{R}\) is the reward function and \(\gamma\) is the discount factor.
Based on a state \(s \in \mathcal{S}\), the agent takes an action \(a \in \mathcal{A}\). With a probability of \(\mathcal{P}(s' \mid s,a)\), the agent transitions to the next state \(s' \in \mathcal{S}\).
After transitioning to state \(s'\), the agent receives the reward given by \(\mathcal{R}(s, a, s')\).\\

\paragraph{\texorpdfstring{\acrfull{dql}}{Deep Q Learning (DQL)}}
\acrshort{dql} performs Q-Learning \citep{watkins_learning_1989} by leveraging a neural network (parameterized with \(\theta\)) to approximate the (optimal) action-value function \(Q^*(s,a)\), so that \(Q(s,a,\theta) \approx Q^*(s,a)\) \citep{mnih_playing_2013}.
Like the Q-Learning algorithm, it is a model-free and off-policy algorithm.

\paragraph{\texorpdfstring{\acrfull{a2c}}{Advantage Actor-Critic (A2C)}}
\acrshort{a2c} is an advanced Actor-Critic method, where an advantage function \(A(a_t, s_t)\) acts as the critic \citep{mnih_asynchronous_2016}. That is, \(A(a_t, s_t) = Q(a_t, s_t) - V(s_t)\) where \(Q(a_t, s_t)\) is the action-value function and \(V(s_t)\) is the state-value function.
\(A(a_t, s_t)\) describes the advantage of taking action \(a_t\) in state \(s_t\).

\paragraph{\texorpdfstring{\acrfull{ppo}}{Proximal Policy Optimization}}
\acrshort{ppo} is an Actor-Critic method that is more stable than other algorithms such as \acrshort{a2c} \citep{schulman_proximal_2017}.
\acrshort{ppo} uses a clipped "surrogate" objective function \(L^{\text{CLIP}}\): {
\begin{equation} \label{eq:l_clip}
    L^{\text{CLIP}}(\theta) = \hat{\mathbb{E}}_t \left[\min \left(r_t(\theta) \hat{A}_t, \text{clip}\left(r_t(\theta), 1 - \epsilon, 1 + \epsilon\right)\hat{A}_t\right)\right]
\end{equation} } 
with \(\epsilon\) being a hyperparameter (originally \(\epsilon = 0.2\)).
The first term \(r_t(\theta) \hat{A}_t\) corresponds to the \(L^{\text{CPI}}\).
\(\text{CPI}\) stands for \acrlong{cpi}.
The \(\text{clip}\) function clips the probability ratio \(r_t(\theta)\) by \(1 - \epsilon, 1 + \epsilon\).
By doing so, a too large policy update is avoided.

%% file: sections/3_methodology.tex
\section{Methodology}

The work was conducted in multiple phases (Appendix \ref{sec:project_structure}).
For each phase, options were explored (divergence) and then the best option was chosen (convergence).

\subsection{The Pure \texorpdfstring{\acrshort{got}}{GoT} Framework}

Instead of relying on the \acrlong{got} framework of \citet{besta_graph_2024}, an adapted one was built on its core concepts (Appendix \ref{sec:pure_graph_of_thoughts}).
Despite the original framework having an extensible \acrshort{api}, its inherent structure does not suit the needs of this automation approach.
Our adapted \acrshort{got} framework aims for simplicity and expressiveness, while enabling processing-independent task definitions and iterative processing.

\paragraph{Task}
A task is defined as a comprehensive piece of work. A task can be described by a set of possible operations \(\mathcal{O}\).

\paragraph{Operation}
An operation \(o: S^{n} \rightarrow S^{m}\) can be seen as a function that transforms incoming states to outgoing states (with \(n\) input thoughts and \(m\) output thoughts, respectively).
An operation can be classified in two variants.
1) A prompt operation is executed by prompting a language model.
2) An execution operation is a programmatically defined function that is evaluated computationally.

\paragraph{Score Operation}
A score operation \(o_s: \mathbb{R} \times S \times S \rightarrow \mathbb{R}\) is applied to get the score of a thought produced by an operation. As input, the score operation receives the cumulative score and both the previous and the current state. The score is represented as a real number.
Identical to the common operation, the score operation can either be a prompt operation or a computed execution operation.

\paragraph{Complexity}
We define complexity as a measurement of difficulty of a task's instance \footnote{In contrary to other definitions such as computational complexity.}.
It is represented by a numerical value that is specific to a task.
The complexity is directly dependent on the input (e.g. number of elements in a list to sum).

\paragraph{\texorpdfstring{\acrlong{goo}}{Graph of Operations}}

\begin{figure}[t]
    \centering
    \subfloat[\acrshort{goo}\label{fig:example_goo}]{%
        \includegraphics[height=0.18\textheight]{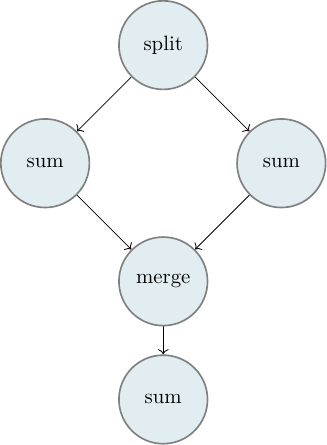}}%
    \qquad
    \subfloat[\acrshort{got}\label{fig:example_got}]{%
        \includegraphics[height=0.18\textheight]{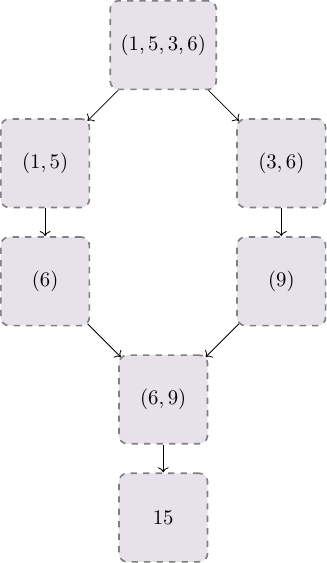}}%
    \qquad
    \subfloat[GoO and GoT\label{fig:example_goo_got_merged}]{%
        \includegraphics[height=0.25\textheight]{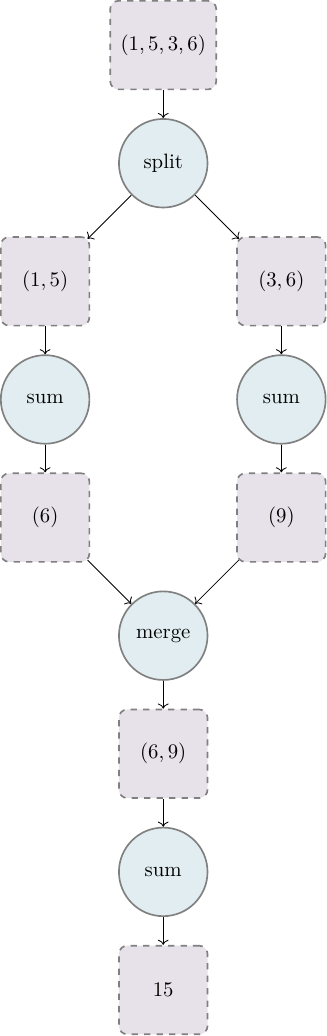}}%
    \caption{Example \acrshort{goo}, \acrshort{got}, and their relationship for the task \task{sum list}.}
    \label{fig:example_goo_got}
\end{figure}

While the reasoning process is modeled in its completeness by \citet{besta_graph_2024}, we specify a \GoO separate from its resulting \GoT and model it with additional constraints (Figure \ref{fig:example_goo}).

A \GoO is a directed, acyclic graph (DAG) \(G \) described by the tuple \((V, E)\) with \(V\) being the set of nodes and \(E\) being the sets of edges.
\(G\) is structured into layers \(L_0, L_1, \ldots, L_n\), where \(n \in \mathbb{N}_0\) and each layer \(L_i\) contains a subset of the nodes, such that \(V = \bigcup_{i=0}^n L_i\) and \(L_i \cap L_j = \emptyset\) for all \(i \neq j\). The index of the layer corresponds to the distance of the nodes to the source node.
An edge can solely be formed between nodes of directly succeeding layers. That is, for each edge \((u, v) \in E\), if \(u \in L_i\) and \(v \in L_j\), then \(|i-j| = 1\).

\(G\) has exactly one source and exactly one sink, so that \(|L_0| = 1 \land |L_n| = 1\). Having these constraints, \(G\) can receive an initial input state and yields a single resulting thought.

By restricting the \GoO in such a way, the graphs can be treated uniformly.

\paragraph{\texorpdfstring{\acrlong{got}}{Graph of Thoughts}}

In this work, the \GoT is separate from the \GoO, but structured similarly (Figure \ref{fig:example_got}).
However, each thought contains a reference to its creating operation, the thought's origin (Figure \ref{fig:example_goo_got_merged}).
An exception is the initial state that is wrapped in a thought without origin.

\FloatBarrier

\subsubsection{Application to \texorpdfstring{\acrshort{rgot}}{RGoT}}
The Pure \acrshort{got} Framework stands independently and is applied as an external component to \acrshort{rgot}.
A custom graph controller allows the step-wise execution of a given graph.

\paragraph{Constraints on Graph Construction} \label{sec:constraints_on_graph_construction}
Due to the graph of operation’s properties, the graph cannot be constructed completely arbitrarily.

Therefore, certain heuristics are employed to obtain a graph with the desired properties.
Both the depth \(d\) and the breadth \(b\) of the graph must not exceed the maximum depth \(d_{\mathit{max}}\) and the maximum breadth \(b_{\mathit{max}}\), respectively.
Since a \GoO must converge towards a single sink, divergence of the graph is not allowed after reaching the divergence cutoff depth \(d_{\neg \mathit{div}}\), which is determined by a divergence cutoff factor \(\mu_{\neg \mathit{div}}\):
\begin{equation} \label{eq:random_graph_construction_divergence_cutoff}
    d_{\neg \mathit{div}} = \lceil d \cdot \mu_{\neg \mathit{div}} \rceil
\end{equation}

Applying these constraints, the construction of a \GoO is guided towards its proper shape.

\subsection{Example Task: \task{sum list}}

In an initial phase, the experiments were conducted on a singular example task named \task{sum list}, with the goal of calculating the sum of a list of single-digit integers.
There are three operations that can be applied in context of the \task{sum list} task.
\begin{itemize}[itemsep=2pt]
    \item \op{sum}: calculate the sum of a given list
    \item \op{split}: split a given list into equally sized sublists
    \item \op{merge}: merge two given lists into a single flat list
\end{itemize}

The operation \op{sum} represents the primary operation for solving the task and features an attached computed score operation to evaluate its produced thoughts.

The task \task{sum list} is chosen as an example task because of its ability to capture the general problem the \acrshort{got} framework is trying to solve, not because the task represents a real world use case.

Mathematical problem solving is challenging for (conventional) \acrshortpl{llm} \citep{ahn_large_2024}.
The task's complexity can be expressed by the length of the list to sum.
The language model's result is only reliable to a certain number of elements in the list.
Thus, the task can only be solved by the input-output strategy up to a certain list length.
Not only must the list's length be reduced by splitting the list into sublists, but the results must be combined to get the total sum of the list's elements. The \acrshort{got} paradigm enables both divergence and convergence of resulting thoughts.

\subsection{Additional Tasks}
In addition to solving our original example task, we evaluate \acrshort{rgot} on tasks similar to those presented by \citet{besta_graph_2024} (Appendix \ref{sec:task_defintions}).

\paragraph{\task{sort list}}
Sorting a list of single-digit integers.
While this is similar in structure as the task \task{sum list}, this task has different properties in terms of complexity reduction.

\paragraph{\task{intersect set}}
Creating the intersection of two sets.
A divide and conquer scheme for set intersection requires a mathematically specific decomposition, which we implemented here statically.

\paragraph{\task{count keywords}}
Counting the number of occurring keywords in a text.
Here, the dataset of \citet{besta_graph_2024} is used with custom sampler logic.

\paragraph{\task{merge documents}}
Merging a list of documents into one concise document, while trying to maximize content retention and minimizing redundancy.
The dataset of \citet{besta_graph_2024} is used.
A noteworthy difference is that we use ROUGE-L for retention scoring and ROUGE-1 for redundancy scoring instead of relying on the language model's judgment \citep{rouge_lin_2004}.

\subsection{\texorpdfstring{\acrshort{llm}}{LLM} Evaluation and Simulation}

To overcome the unwanted elasticity of the training cost due to the sample inefficiency of \acrshort{rl}, the thoughts of the targeted language model are simulated in the \acrshort{rl} environment.
The capability of an \acrshort{llm} to solve a task of a given complexity is modeled as a Bernoulli experiment, where each solution attempt succeeds independently with probability \(P(c): \mathbb{N} \rightarrow \mathbb{R}\), with \(c\) being the complexity of the task.
\(P\) is approximated empirically by repeatedly prompting a language model to solve a corresponding task, yielding the estimate \(\hat{P}(c) = \frac{n_v}{n}\) with \(n_v\) and \(n\) being the number of valid results and the total number of results, respectively.
We use \(n = 100\) per complexity level to fit these estimates.
\(\hat{P}\) then is used to simulate the behavior of the evaluated \acrshort{llm} during training and evaluation.

\paragraph{Selected Language Models}
Three language models of OpenAI are used in this work, of which one is chosen \citep{openai_models_2024} \footnote{The initial experiments were conducted mid 2024 and the then available models were used.
By using those models our experiments are closer to the ones of the original \acrshort{got} paper of \cite{besta_graph_2024}.}:
\begin{itemize}[itemsep=2pt]
    \item \llm{gpt-3.5-turbo-0125}
    \item \llm{gpt-3.5-turbo-1106}
    \item \llm{gpt-4-turbo-2024-04-09}
\end{itemize}

The introduced framework is not limited to OpenAI language models. Any language model can be used by extending the language model \acrshort{api} with a respective interface.

Since our framework and automation approach is fully agnostic to the used language model, using more recent models as simulation targets would not give more insights.
With the improvement of language models, operations may succeed with higher complexities, which ultimately is just a scaling factor, as post-hoc \acrshort{llm} evaluations show (Appendix \ref{sec:post_hoc_llm_evaluation}).

\subsection{\texorpdfstring{\acrshort{rgot}}{RGoT} \texorpdfstring{\acrlong{rl}}{Reinforcement Learning} Configuration}
The agent acts on a custom \acrshort{rl} environment for \acrshort{got} construction. The environment is built on an abstraction layer that introduces more specific spaces such as a boolean space and both categorical and ordinal discrete spaces (Appendix \ref{sec:rl_observation_space_types}).
Experiments with various combinations of network configurations and algorithms, namely \acrshort{dqn}, 
\acrshort{a2c}, 
and \acrshort{ppo}, 
are conducted.
The reward function is defined in a reward shaping process that involves empirical evaluation of different reward functions \citep{wiewiora_reward_2010}.

\subsubsection{Action Space}
The action space \(\mathcal{A}\) is of discrete nature.
The agent acts on a layer basis on the \GoO.
Given the set of operations \(\mathcal{O}\), \(\mathcal{A}\) is defined as
\begin{equation} \label{eq:action_space}
    \mathcal{A} = \{\mathit{stop}, \mathit{backtrack}\} \cup \{\mathit{append}(o) \mid o \in \mathcal{O}\}
\end{equation}
The action \(\mathit{stop}\) signals the end of the \GoO and triggers the ground truth evaluation of the result.
The sink layer of the graph can be removed by playing the \(\mathit{backtrack}\) action.
A layer with nodes of a certain operation can be appended to the graph by playing the action \(\mathit{append}(o)\), where \(o\) is the operation to append.

\subsubsection{Observation Space}
The observation space is a composition of multiple discrete spaces, both categorical and ordinal.

\begin{equation} \label{eq:complete_observation_space}
    \mathcal{S} =
        \mathcal{S}_d \times \mathcal{S}_b \times \mathcal{S}_c \times \mathcal{S}_{\mathit{lc}} \times \mathcal{S}_G \times \mathcal{S}_{\mathit{pa}} \times \mathcal{S}_{\mathit{ps}}
\end{equation}

Each component provides valuable information about the current state (Table \ref{tab:observation_space_components}).

\begin{table}[hb]
    \centering
    \begin{tabular}{l|l|l|l}
        \toprule
        Component & Notation & Type & Values \\
        \midrule
        Depth & \(\mathcal{S}_d\) & Ordinal discrete & \([0, d_{\mathit{max}}]\) \\
        Breadth & \(\mathcal{S}_b\) & Ordinal discrete & \([0, b_{\mathit{max}}]\) \\
        Complexity & \(\mathcal{S}_c\) & Ordinal discrete & \([1, c_{\mathit{max}}]\) \\
        Local complexity & \(\mathcal{S}_{\mathit{lc}}\) & Ordinal discrete & \([1, c_{\mathit{max}}]\) \\
        Graph operations & \(\mathcal{S}_G\) & Multi discrete & \(\vec{v}_G\) \\
        Previous actions & \(\mathcal{S}_{\mathit{pa}}\) & Multi discrete & \(\vec{v}_{\mathit{pa}}\) \\
        Previous score & \(\mathcal{S}_{\mathit{ps}}\) & Optional Boolean & \(\{0, 1, \emptyset\}\) \\
        \bottomrule
    \end{tabular}
    \caption{Observation space components.}
    \label{tab:observation_space_components}
\end{table}

\paragraph{Graph Representation}
The \GoO is represented in a reduced form based on the graph's layer.
Each layer is represented by the operation of the layer's nodes.
In this scenario, the benefit of reducing the graph's representation from two dimensions to a single one outweighs the drawback of the lost flexibility.
With a lower complexity of the inputs, the agent can learn the impact of the graph structure faster.

%% file: sections/4_results.tex
\clearpage

\section{Results}

\subsection{\texorpdfstring{\acrshort{llm}}{LLM} Capability Evaluation}

Initially, three \acrshortpl{llm} were evaluated on the operation \op{sum} of the task \task{sum list} (Appendix \ref{sec:additional_llm_evaluation}).
Because of its good price-performance ratio, the model \llm{gpt-3.5-turbo-0125} was chosen to be the target model and its \(\hat{P}\) was used for the simulation.

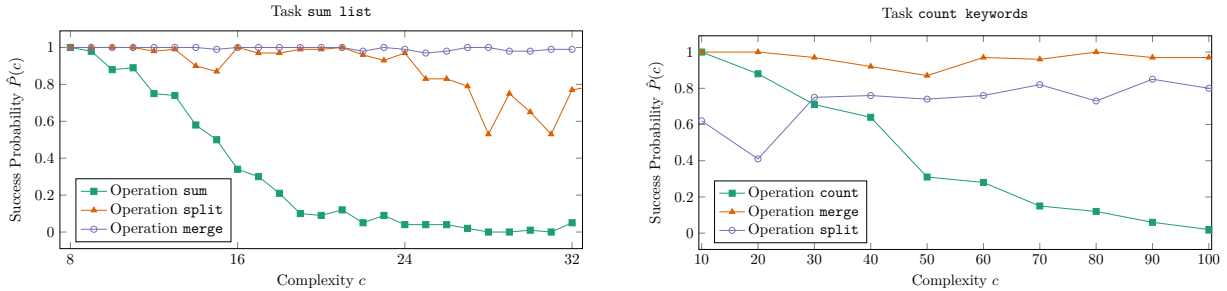
\begin{figure}[hb]
    \centering
    \subfloat{%
        \resizebox{0.48\linewidth}{!}{
            \input{results/llm_capability_evaluation/sum_list/sum_list_all_ops}
        }
    }
    \hfill
    \subfloat{%
        \resizebox{0.48\linewidth}{!}{
            \input{results/llm_capability_evaluation/count_keywords/count_keywords_all_ops}
        }
    }
    \caption{Operation success probability \(\hat{P}(c)\) as a function of complexity \(c\) for the operations of the tasks \task{sum list} and \task{count keywords}, evaluated with \llm{gpt-3.5-turbo-0125}.}
    \label{fig:llm_capability_evaluation}
\end{figure}

The pattern of decreasing probability of success with an increase of complexity is consistent across all critical operations of the example tasks (Figure \ref{fig:llm_capability_evaluation}, Appendix \ref{sec:llm_capability_evaluation_results}).
This property validates and reinforces the use of a prompting paradigm that is capable of divergence and convergence.

\subsection{\texorpdfstring{\acrlong{rgot}}{Reinforced Graph of Thoughts} Agents}

\begin{figure}[t]
    \centering
    \subfloat[\task{sum list}]{%
        \includegraphics[width=0.45\linewidth]{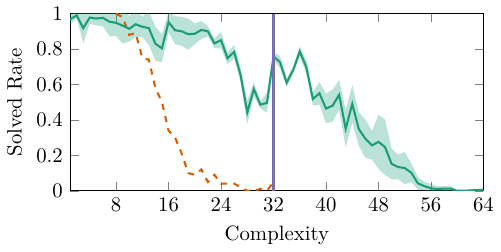}
    }\hfill
    \subfloat[\task{sort list}]{%
        \includegraphics[width=0.45\linewidth]{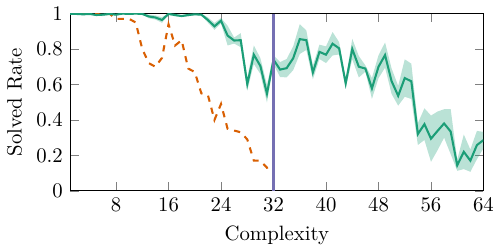}
    }\hfill
    \subfloat[\task{count keywords}]{%
        \includegraphics[width=0.45\linewidth]{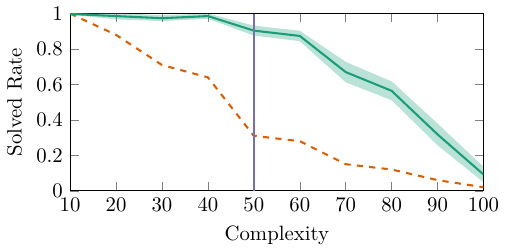}
    }\hfill
    \subfloat[\task{intersect set}]{%
        \includegraphics[width=0.45\linewidth]{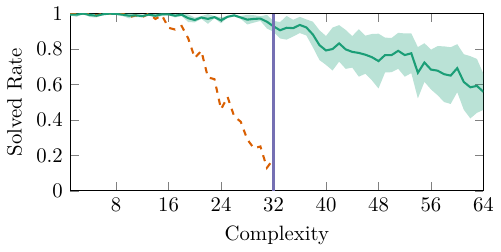}
    }\hfill
    \subfloat[\task{merge docs}]{%
        \includegraphics[width=0.45\linewidth]{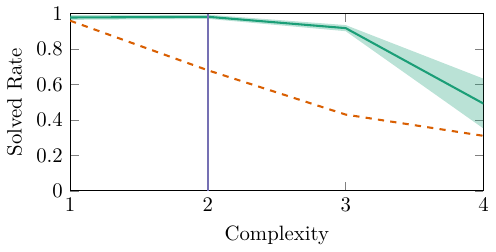}
    }\hfill
    \subfloat[]{%
        \begin{tikzpicture}
            \begin{axis}[
                hide axis,
                width=0.4\textwidth,
                height=0.4\textwidth,
                legend columns=1,
                legend style={draw=none, anchor=east, at={(-0.5,0.5)}},
                legend cell align={left},
                xmin=0, xmax=1, ymin=0, ymax=1,
            ]
                \addplot [draw=none, forget plot] coordinates {(0,0)};
        
                \addlegendimage{color=category a, line width=1.5pt}
                \addlegendentry{\acrshort{rgot} Agent (mean $\pm$ std)}
                \addlegendimage{color=category b, line width=1.5pt, dashed}
                \addlegendentry{Input-Output (IO) baseline}
                \addlegendimage{color=category c, line width=1.5pt}
                \addlegendentry{Training Front (inclusive)}
            \end{axis}
        \end{tikzpicture}%
    }\hfill
    \caption{Solved rate per complexity for all five tasks. \acrshort{rgot} agents (teal) are compared against the IO baselines (orange). The vertical line marks the training front, results to its right are out-of-distribution.}
    \label{fig:auto_got_solved_rates}
\end{figure}
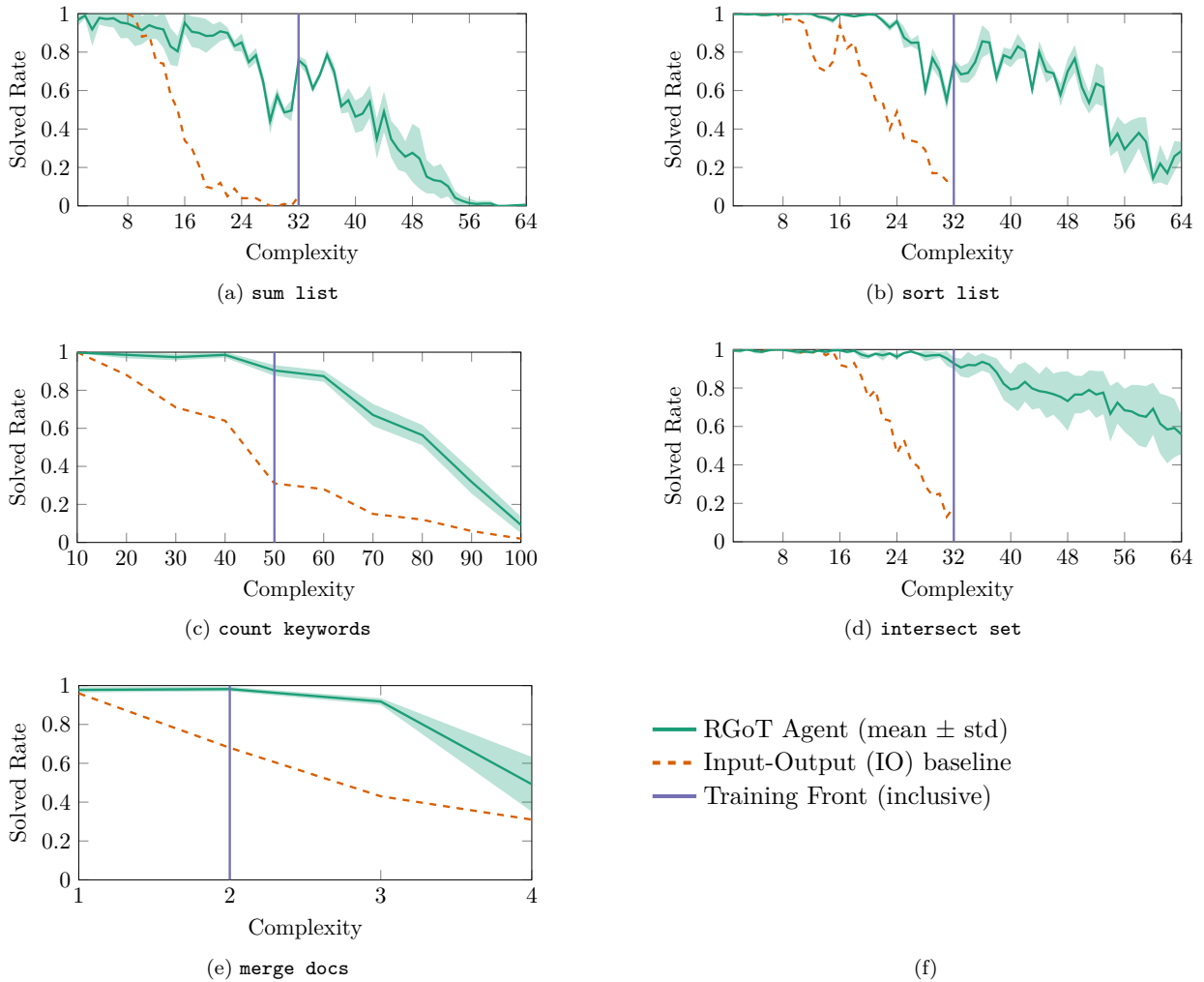

We employ simple \acrshort{ppo} agents with \(2\) layers of \(64\) units (for both actor and critic network) (Table \ref{tab:final_configuration}, Appendix \ref{sec:ablation_summary}).
The agents achieve solid results across all tasks (Figure \ref{fig:auto_got_solved_rates}).
After \(2^{18}\) training timesteps, the agents are able to solve \(0.9399\pm0.0533\) of the training complexities on average across all tasks.
Moreover, the agents can solve out-of-distribution complexities, resulting in a total average solved rate\footnote{Mean solved rate per task computed over all evaluated complexity levels, then averaged across the five tasks} of \(0.7512\pm0.1089\).
In comparison to the input-output (\acrshort{io}) prompting strategy, the agent can solve tasks of higher complexities more reliably.

\begin{table}[h]
    \centering
    \subfloat[PPO hyperparameters.]{
        \begin{tabular}{lp{5cm}}
            \toprule
            Hyperparameter & Value \\
            \midrule
            Clip range & linear schedule \(0.15 \to 0.30\) \\
            Learning rate & linear decay \( 5 \times 10^{-4} \to 1\times10^{-6}\),\newline\(10\%\) warmup \\
            Update epochs \(K\) & \(8\) \\
            Entropy coefficient & \(0.03\) \\
            Policy / value network & \([64, 64]\) / \([64, 64]\) \\
            Total timesteps & \(2^{18}\) \\
            Parallel environments   & \(8\) \\
            \bottomrule
        \end{tabular}
    }
    \hfill
    \subfloat[Environment parameters.]{
        \begin{tabular}{ll}
            \toprule
            Parameter & Value \\
            \midrule
            Max steps & \(20\) \\
            Max depth \(d_{\mathit{max}}\) & \(8\) \\
            Max breadth \(b_{\mathit{max}}\) & \(8\) \\
            Divergence cutoff factor \(d_{\neg \mathit{div}}\) & \(0.5\) \\
            Max operations & \(32\) \\
            \bottomrule
        \end{tabular}
    }
    \caption{Final \acrshort{ppo} and environment configuration.}
    \label{tab:final_configuration}
\end{table}

%% file: results/llm_capability_evaluation/sum_list/sum_list_all_ops.tex
\begin{tikzpicture}
    \begin{axis}[
        title={Task \task{sum list}},
        xtick distance=8,
        ytick distance=0.2,
        ymin=0, ymax=1,
        xmax=32,
        enlarge x limits={abs=0.5},
        enlarge y limits={abs=0.09},
        scale only axis,
        xlabel={Complexity \(c\)},
        ylabel={Success Probability \(\hat{P}(c)\)},
        height=5cm,
        width=12cm,
        legend pos=south west,
        legend cell align={left},
        cycle list name=mark categories,
    ]
        \addplot table [x=cardinality, y=probability, col sep=comma]
            {results/llm_capability_evaluation/sum_list/op_sum/gpt35turbo0125.csv};
        \addlegendentry{Operation \op{sum}}
        
        \addplot table [x=cardinality, y=probability, col sep=comma]
            {results/llm_capability_evaluation/sum_list/op_split/gpt35turbo0125.csv};
        \addlegendentry{Operation \op{split}}

        \addplot table [x=cardinality, y=probability, col sep=comma]
            {results/llm_capability_evaluation/sum_list/op_merge/gpt35turbo0125.csv};
        \addlegendentry{Operation \op{merge}}
    \end{axis}
\end{tikzpicture}

%% file: results/llm_capability_evaluation/count_keywords/count_keywords_all_ops.tex
\begin{tikzpicture}
    \begin{axis}[
        title={Task \task{count keywords}},
        xtick distance=10,
        ytick distance=0.2,
        ymin=0, ymax=1,
        xmax=100,
        enlarge x limits={abs=0.5},
        enlarge y limits={abs=0.09},
        scale only axis,
        xlabel={Complexity \(c\)},
        ylabel={Success Probability \(\hat{P}(c)\)},
        height=5cm,
        width=12cm,
        legend pos=south west,
        legend cell align={left},
        cycle list name=mark categories,
    ]
        \addplot table [x=cardinality, y=probability, col sep=comma]
            {results/llm_capability_evaluation/count_keywords/op_count/gpt35turbo0125.csv};
        \addlegendentry{Operation \op{count}}

        \addplot table [x=cardinality, y=probability, col sep=comma]
            {results/llm_capability_evaluation/count_keywords/op_merge/gpt35turbo0125.csv};
        \addlegendentry{Operation \op{merge}}

        \addplot table [x=cardinality, y=probability, col sep=comma]
            {results/llm_capability_evaluation/count_keywords/op_split/gpt35turbo0125.csv};
        \addlegendentry{Operation \op{split}}
    \end{axis}
\end{tikzpicture}

%% file: sections/5_discussion.tex
\FloatBarrier
\section{Discussion}

\subsection{Agent Performance}
There is a generalization of the agent that can be observed.
The primary indicator is the number of operations that increases with a higher complexity (Appendix \ref{sec:agent_evaluation_results}), even after the \gls{tf} (Figure \ref{fig:auto_got_solved_rates}).
Despite the absence of the complexities in the training period, the agent decides to increase the branching to solve the task.
This leads to the agent's approach showing a significantly higher solved rate in comparison to the baseline.

\paragraph{Task Complexities}
One of the key mechanisms for the agent to learn is when to branch instead of applying the primary operation directly.
Since the range of complexity levels seen during training is limited, the agent has few opportunities to observe the transition points where decomposition becomes necessary.
This may negatively impact generalization, as the agent has limited experience with the branching conditions that dominate at higher complexities.

\paragraph{Variance of Agent Performance}
Some agents show a higher variance in their performance across different seeds (Figure \ref{fig:auto_got_solved_rates}).
A possible explanation is a high learning rate sensitivity: as different tasks have different reward landscapes, the applied learning rate schedule may not be optimal for all of them.
Another difficulty may be the episode-level noise of the language model simulation. Since the simulation is stochastic, the agent may receive mixed signals for the same graph structure.
Furthermore, the sparse reward signal beyond the training front amplifies seed sensitivity: successful episodes become rare at high complexity, so early differences in sampled trajectories can cause the policy to commit to divergent strategies across seeds.
This effect lies in the nature of \acrshort{ppo} being on-policy: unlike off-policy methods, \acrshort{ppo} cannot revisit past successful trajectories and therefore early unlucky seeds cannot recover the missed signal.
Consequently, variance is most pronounced in the out-of-distribution region, where the reward signal is weakest and recovery from early suboptimal commitments is least likely.

\subsection{Accessibility and Adaptability}
By automatically generating a \GoO, the \acrshort{got} prompting paradigm is more accessible and requires fewer manual steps and knowledge.
More importantly, the agent adapts the operations with respect to the task's complexity, making the paradigm usable to a broader range of tasks.

\subsection{Usability}
A key property of \acrshort{rgot} is it being completely language model agnostic.
With respect to a task, a language model's behavior can be simulated by probability distributions of the success of the task's operations.
An agent can then be trained on the desired model's simulation, offering a cost-effective way of training.
When using larger and more capable models only shifts the success range of a task's operation to higher complexities (Appendix \ref{sec:post_hoc_llm_evaluation}), one could use smaller models to achieve the same task by using our \acrshort{rl} approach.

\subsection{Potential Applications}
The proposed method is potentially applicable to various tasks that can be solved by a language model, such as merging documents, counting keywords in a text or summarizing large documents.

In the future, it could be possible to create an automated pipeline to train a \RL agent that can solve a defined task.
Additionally, a library of trained agents could be provided for a set of common tasks of a certain domain, so that users can employ the \acrshort{got} paradigm without additional effort.
Moreover, the \acrshort{rgot} approach could be used in combination with \acrfull{rag} \citep{lewis_retrieval_2020} to both preprocess and postprocess information.

A related direction is \acrfull{kgot} \cite{besta_kgot_2025}, which dynamically constructs a knowledge graph to represent the evolving state of an AI assistant task.
Since \acrshort{kgot} structures externally retrieved knowledge while \acrshort{rgot} structures the reasoning topology itself, the two paradigms are complementary and could be combined - for example, an \acrshort{rgot} agent could adaptively govern the operation-level structure within a \acrshort{kgot} pipeline.

\subsection{Conclusion}
To the best of our knowledge, this is the first work to present an automated approach for constructing a \GoT by automatically generating and traversing a \GoO without relying on a language model's ability to plan or to know the properties of a presented task.
The empirical results indicate that, under certain constraints, it is possible to employ an automated approach to the \acrlong{got} prompting paradigm by using a \RL agent.

%% file: sections/6_appendix.tex
\input{sections/appendix/project_structure}
\clearpage
\input{sections/appendix/pure_graph_of_thoughts}
\clearpage
\input{sections/appendix/technical_details}
\clearpage
\input{sections/appendix/rl_observation_spaces}
\clearpage
\input{sections/appendix/reward_function}
\clearpage
\input{sections/appendix/task_definitions}
\clearpage
\input{sections/appendix/llm_capability_evaluation_results}
\clearpage
\input{sections/appendix/agent_evaluation_results}
\clearpage
\input{sections/appendix/ablation_summary}
\clearpage
\newpage

\vfill\eject

%% file: sections/appendix/project_structure.tex
\section{Project Structure}
\label{sec:project_structure}

\begin{figure}[hb]
    \centering
    \resizebox{\linewidth}{!}{
        \input{diagrams/project_structure}
    }
    \caption{Project Structure}
    \label{fig:project_structure}
\end{figure}
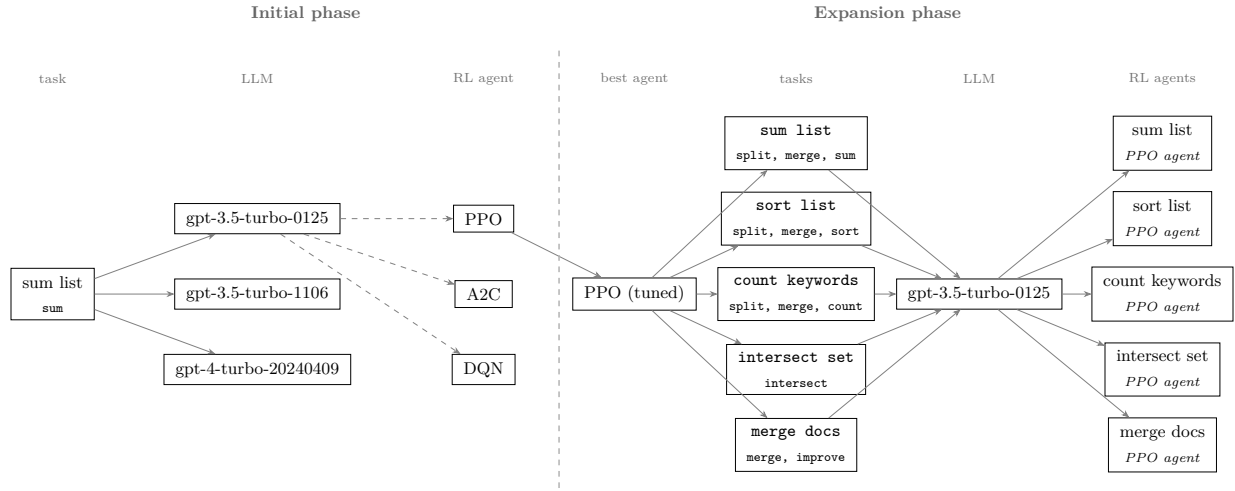

The project consists of two phases (Figure \ref{fig:project_structure}).
\paragraph{Initial phase}
In a first phase, the operation \op{sum} of the task \task{sum list} was evaluated on different language models, of which then \llm{gpt-3.5-turbo-0125} was selected.
Different \acrshort{rl} algorithms were then evaluated with a variety of configurations, of which one \acrshort{ppo} was selected for further tuning.
This provided a foundation for further experiments.

\paragraph{Expansion phase}
Since a single example task is not representative, it was decided to extend the set of tasks with tasks similar to the ones shown by \citet{besta_graph_2024}.
Moreover, all operations were evaluated on \llm{gpt-3.5-turbo-0125}.
Then, for each task a \acrshort{ppo} agent was trained.

%% file: diagrams/project_structure.tex
\tikzset{
  nd/.style={draw, thin, align=center, inner xsep=6pt, inner ysep=4pt, font=\small},
  arr/.style={-{Stealth[length=4pt,width=3pt]}, thin, draw=black!50},
  sel/.style={-{Stealth[length=4pt,width=3pt]}, thin, draw=black!50, dashed},
  lbl/.style={font=\scriptsize, text=black!55},
}
\begin{tikzpicture}
\node[nd] (task) at (0, 0) {sum list\\[2pt]{\scriptsize\texttt{sum}}};
\node[nd] (m1) at (3.8,  1.4) {gpt-3.5-turbo-0125};
\node[nd] (m2) at (3.8,  0.0) {gpt-3.5-turbo-1106};
\node[nd] (m3) at (3.8, -1.4) {gpt-4-turbo-20240409};
\node[nd] (ppo) at (8.0,  1.4) {PPO};
\node[nd] (a2c) at (8.0,  0.0) {A2C};
\node[nd] (dqn) at (8.0, -1.4) {DQN};
\node[nd] (ppot) at (10.8, 0.0) {PPO (tuned)};
\node[nd] (ts1) at (13.8,  2.8) {\task{sum list}\\[2pt]{\scriptsize\texttt{split, merge, sum}}};
\node[nd] (ts2) at (13.8,  1.4) {\task{sort list}\\[2pt]{\scriptsize\texttt{split, merge, sort}}};
\node[nd] (ts3) at (13.8,  0.0) {\task{count keywords}\\[2pt]{\scriptsize\texttt{split, merge, count}}};
\node[nd] (ts4) at (13.8, -1.4) {\task{intersect set}\\[2pt]{\scriptsize\texttt{intersect}}};
\node[nd] (ts5) at (13.8, -2.8) {\task{merge docs}\\[2pt]{\scriptsize\texttt{merge, improve}}};
\node[nd] (mfix) at (17.2, 0.0) {gpt-3.5-turbo-0125};
\node[nd] (ag1) at (20.6,  2.8) {sum list\\[2pt]{\scriptsize\textit{PPO agent}}};
\node[nd] (ag2) at (20.6,  1.4) {sort list\\[2pt]{\scriptsize\textit{PPO agent}}};
\node[nd] (ag3) at (20.6,  0.0) {count keywords\\[2pt]{\scriptsize\textit{PPO agent}}};
\node[nd] (ag4) at (20.6, -1.4) {intersect set\\[2pt]{\scriptsize\textit{PPO agent}}};
\node[nd] (ag5) at (20.6, -2.8) {merge docs\\[2pt]{\scriptsize\textit{PPO agent}}};
\draw[arr] (task) -- (m1);
\draw[arr] (task) -- (m2);
\draw[arr] (task) -- (m3);
\draw[sel] (m1) -- (ppo);
\draw[sel] (m1) -- (a2c);
\draw[sel] (m1) -- (dqn);
\draw[arr] (ppo) -- (ppot);
\draw[arr] (ppot) -- (ts1);
\draw[arr] (ppot) -- (ts2);
\draw[arr] (ppot) -- (ts3);
\draw[arr] (ppot) -- (ts4);
\draw[arr] (ppot) -- (ts5);
\draw[arr] (ts1) -- (mfix);
\draw[arr] (ts2) -- (mfix);
\draw[arr] (ts3) -- (mfix);
\draw[arr] (ts4) -- (mfix);
\draw[arr] (ts5) -- (mfix);
\draw[arr] (mfix) -- (ag1);
\draw[arr] (mfix) -- (ag2);
\draw[arr] (mfix) -- (ag3);
\draw[arr] (mfix) -- (ag4);
\draw[arr] (mfix) -- (ag5);
\node[lbl] at (0,    4.0) {task};
\node[lbl] at (3.8,  4.0) {LLM};
\node[lbl] at (8.0,  4.0) {RL agent};
\node[lbl] at (10.8, 4.0) {best agent};
\node[lbl] at (13.8, 4.0) {tasks};
\node[lbl] at (17.2, 4.0) {LLM};
\node[lbl] at (20.6, 4.0) {RL agents};

\draw[black!30, dashed, thick] (9.4, -3.6) -- (9.4, 4.6);
\node[font=\small\bfseries, text=black!60] at (4.7,  5.2) {Initial phase};
\node[font=\small\bfseries, text=black!60] at (15.5, 5.2) {Expansion phase};

\end{tikzpicture}

%% file: sections/appendix/pure_graph_of_thoughts.tex
\section{Pure Graph of Thoughts}
\label{sec:pure_graph_of_thoughts}

Our Pure Graph of Thoughts framework addresses the absence of some functionalities of the original \acrshort{got} framework and represents the foundation for the \acrshort{rgot} framework.

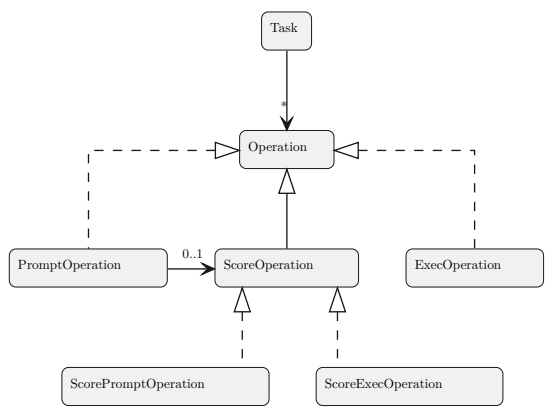
\begin{figure}[hb]
    \centering
    \resizebox{0.45\textwidth}{!}{
        \input{./diagrams/pure_got_types}
    }
    \caption{Pure Graph of Thoughts Types}
    \label{fig:pure_got_types}
\end{figure}

As a user-facing \acrshort{api}, operations can be defined in a declarative way over a typed and validated data structure (\acrshort{dsl}) (Figure \ref{fig:pure_got_types}).
This allows the definition of operations and tasks decoupled to the underlying processing.
There is a strict distinction between a prompt operation executed by a language model and a code execution operation.
To simplify parsing logic and to ensure consistent results, the \acrshort{json} format is used for communication with the language model.
The scoring is now part of an operation involving a prompt, rather than being a standalone operation that can be added arbitrarily. While this simplifies the automation process, it restricts the user's possibility of adding a validation operation.
While the framework is crafted with automation in mind, it allows the manual construction of \GoOs.
Operations and thoughts are represented independently of their graph structure.

\vspace{0.25cm}

\noindent
The related code can be found in the following repository: 
\ifblind
<blinded for review>
\else
\url{https://github.com/mriesen/pure-graph-of-thoughts}
\fi

%% file: diagrams/pure_got_types.tex
\definecolor{plantucolor0000}{RGB}{241,241,241}
\definecolor{plantucolor0001}{RGB}{24,24,24}
\definecolor{plantucolor0002}{RGB}{0,0,0}
\begin{tikzpicture}[yscale=-1
    ,pstyle0/.style={color=plantucolor0001,fill=plantucolor0000,line width=0.5pt}
    ,pstyle1/.style={color=plantucolor0001,line width=1.0pt}
    ,pstyle2/.style={color=plantucolor0001,fill=plantucolor0001,line width=1.0pt}
    ,pstyle3/.style={color=plantucolor0001,line width=1.0pt,dash pattern=on 7.0pt off 7.0pt}
    ]
    \draw[pstyle0] (196.5pt,12pt) arc (180:270:5pt) -- (201.5pt,7pt) -- (229.0459pt,7pt) arc (270:360:5pt) -- (234.0459pt,12pt) -- (234.0459pt,30.6211pt) arc (0:90:5pt) -- (229.0459pt,35.6211pt) -- (201.5pt,35.6211pt) arc (90:180:5pt) -- (196.5pt,30.6211pt) -- cycle;
    \node at (199.5pt,12pt)[below right,color=black]{Task};
    \draw[pstyle0] (180pt,101pt) arc (180:270:5pt) -- (185pt,96pt) -- (245.8905pt,96pt) arc (270:360:5pt) -- (250.8905pt,101pt) -- (250.8905pt,119.6211pt) arc (0:90:5pt) -- (245.8905pt,124.6211pt) -- (185pt,124.6211pt) arc (90:180:5pt) -- (180pt,119.6211pt) -- cycle;
    \node at (183pt,101pt)[below right,color=black]{Operation};
    \draw[pstyle0] (305pt,190pt) arc (180:270:5pt) -- (310pt,185pt) -- (403.3239pt,185pt) arc (270:360:5pt) -- (408.3239pt,190pt) -- (408.3239pt,208.6211pt) arc (0:90:5pt) -- (403.3239pt,213.6211pt) -- (310pt,213.6211pt) arc (90:180:5pt) -- (305pt,208.6211pt) -- cycle;
    \node at (308pt,190pt)[below right,color=black]{ExecOperation};
    \draw[pstyle0] (161.5pt,190pt) arc (180:270:5pt) -- (166.5pt,185pt) -- (264.3406pt,185pt) arc (270:360:5pt) -- (269.3406pt,190pt) -- (269.3406pt,208.6211pt) arc (0:90:5pt) -- (264.3406pt,213.6211pt) -- (166.5pt,213.6211pt) arc (90:180:5pt) -- (161.5pt,208.6211pt) -- cycle;
    \node at (164.5pt,190pt)[below right,color=black]{ScoreOperation};
    \draw[pstyle0] (7pt,190pt) arc (180:270:5pt) -- (12pt,185pt) -- (120.6808pt,185pt) arc (270:360:5pt) -- (125.6808pt,190pt) -- (125.6808pt,208.6211pt) arc (0:90:5pt) -- (120.6808pt,213.6211pt) -- (12pt,213.6211pt) arc (90:180:5pt) -- (7pt,208.6211pt) -- cycle;
    \node at (10pt,190pt)[below right,color=black]{PromptOperation};
    \draw[pstyle0] (46.5pt,279pt) arc (180:270:5pt) -- (51.5pt,274pt) -- (197.1309pt,274pt) arc (270:360:5pt) -- (202.1309pt,279pt) -- (202.1309pt,297.6211pt) arc (0:90:5pt) -- (197.1309pt,302.6211pt) -- (51.5pt,302.6211pt) arc (90:180:5pt) -- (46.5pt,297.6211pt) -- cycle;
    \node at (49.5pt,279pt)[below right,color=black]{ScorePromptOperation};
    \draw[pstyle0] (237.5pt,279pt) arc (180:270:5pt) -- (242.5pt,274pt) -- (372.774pt,274pt) arc (270:360:5pt) -- (377.774pt,279pt) -- (377.774pt,297.6211pt) arc (0:90:5pt) -- (372.774pt,302.6211pt) -- (242.5pt,302.6211pt) arc (90:180:5pt) -- (237.5pt,297.6211pt) -- cycle;
    \node at (240.5pt,279pt)[below right,color=black]{ScoreExecOperation};
    \draw[pstyle1] (215.5pt,36.2pt) ..controls (215.5pt,52.69pt) and (215.5pt,73.28pt) .. (215.5pt,89.78pt);
    \draw[pstyle2] (215.5pt,95.78pt) -- (219.5pt,86.78pt) -- (215.5pt,90.78pt) -- (211.5pt,86.78pt) -- (215.5pt,95.78pt) -- cycle;
    \node at (207.3603pt,70.6272pt)[below right,color=black]{*};
    \draw[pstyle3] (161.81pt,111pt) ..controls (117.73pt,111pt) and (66.5pt,111pt) .. (66.5pt,111pt) ..controls (66.5pt,111pt) and (66.5pt,159.75pt) .. (66.5pt,184.89pt);
    \draw[pstyle1] (179.81pt,111pt) -- (161.81pt,105pt) -- (161.81pt,117pt) -- (179.81pt,111pt) -- cycle;
    \draw[pstyle1] (215.5pt,143.2pt) ..controls (215.5pt,159.69pt) and (215.5pt,168.28pt) .. (215.5pt,184.78pt);
    \draw[pstyle1] (215.5pt,125.2pt) -- (209.5pt,143.2pt) -- (221.5pt,143.2pt) -- (215.5pt,125.2pt) -- cycle;
    \draw[pstyle3] (269.08pt,111pt) ..controls (310.85pt,111pt) and (356.5pt,111pt) .. (356.5pt,111pt) ..controls (356.5pt,111pt) and (356.5pt,159.75pt) .. (356.5pt,184.89pt);
    \draw[pstyle1] (251.08pt,111pt) -- (269.08pt,117pt) -- (269.08pt,105pt) -- (251.08pt,111pt) -- cycle;
    \draw[pstyle1] (126.25pt,200pt) ..controls (137.74pt,200pt) and (143.73pt,200pt) .. (155.05pt,200pt);
    \draw[pstyle2] (161.05pt,200pt) -- (152.05pt,196pt) -- (156.05pt,200pt) -- (152.05pt,204pt) -- (161.05pt,200pt) -- cycle;
    \node at (133.6061pt,181.7837pt)[below right,color=black]{0..1};
    \draw[pstyle3] (182pt,232.2pt) ..controls (182pt,248.69pt) and (182pt,257.28pt) .. (182pt,273.78pt);
    \draw[pstyle1] (182pt,214.2pt) -- (176pt,232.2pt) -- (188pt,232.2pt) -- (182pt,214.2pt) -- cycle;
    \draw[pstyle3] (253.5pt,232.2pt) ..controls (253.5pt,248.69pt) and (253.5pt,257.28pt) .. (253.5pt,273.78pt);
    \draw[pstyle1] (253.5pt,214.2pt) -- (247.5pt,232.2pt) -- (259.5pt,232.2pt) -- (253.5pt,214.2pt) -- cycle;
\end{tikzpicture}

%% file: sections/appendix/technical_details.tex
\section{Technical Details}
\label{sec:technical_details}

\subsection{Randomness}
In the context of this work, whenever reference is made to randomly generated numbers or structures, it is always referred to pseudo-random numbers generated by a \acrfull{prng}.
More specifically, Mersenne Twister \citep{matsumoto_mersenne_1998} (Python default) and PCG \citep{oneill_pcg_2014} (NumPy default) are used for CPU-based pseudo-random number generation, whereas the pseudo-randomness on GPU devices is sourced from the Philox algorithm \citep{salmon_parallel_2011} (PyTorch default).

To ensure reproducibility, the \acrshortpl{prng} are used with a fixed seed.
For the language model evaluation, the seed \(0\) is used.
The training and evaluation of the \acrshort{rl} agents incorporates multiple seeds, of which the seed set \(\{0, 8, 16, 24, 32\}\) is specified.
The \acrshort{rl} training uses \(8\) vectorized environments, where each of the environments uses its own seed derived by addition of the specified seed and the rank \([0..8[\).
This leads to the use of all seeds in \([0..32[\) (with some exceptions \footnote{The seeding of the language model simulators is decoupled by adding \(10^5\) to it.}).

\subsection{Technological Overview}
The source code of this work is produced in Python, using the defined \acrshortpl{api} and libraries (Table \ref{tab:technological_overview}) \citep{python_web_2026}.

For the \acrshort{rl} environment definition, the \Gls{gym} \acrshort{api} is used, which is a widely used standard \citep{gymnasium_web_2026}.

Where possible, the \acrshort{rl} algorithm implementations of \Gls{sb3} are used. \Gls{sb3} provides reliable implementations of common \acrshort{rl} algorithms in \Gls{pytorch} \citep{raffin_stablebaselines3_2021}.

\begin{table}[hb]
    \centering
    \begin{tabular}{ll}
        \toprule
        Name & Usage \\
        \midrule
        Python & Programming language \\
        \Gls{gym} & \acrshort{api} standard for \acrshort{rl} environment definition \\
        \Gls{sb3} & Implementations of \acrshort{rl} algorithms \\
        \bottomrule
    \end{tabular}
    \caption{Technological Overview}
    \label{tab:technological_overview}
\end{table}

%% file: sections/appendix/rl_observation_spaces.tex
\section{\texorpdfstring{\acrshort{rl}}{RL} Observation Space Types}
\label{sec:rl_observation_space_types}

Instead of forming the observation space directly out of the spaces defined by \Gls{gym}, an abstraction layer is put on top.
This allows more specific definitions of spaces (Table \ref{tab:observation_space_types}).

\paragraph{Categorical vs ordinal discrete spaces}
When it  comes to encoding categorical data, one-hot encoding is chosen over a numeric encoding in the form of integers or continuous values. This applies for both the actions and the observations of the reinforcement learning environment. Numeric encoding is more space efficient, but introduces an ordinal relationship between the categories.
In the case of independent, unordered categories, such ordinal relationships increase the complexity for the underlying model and may have an impact on the model’s performance.
While the one-hot encoding is less space efficient than numeric encoding, it does not introduce an ordinal relationship between the categories.
Thus, it represents the categories better than numeric encoding, which may be the reason it is widely used \citep{hancock_survey_categorical_2020}.

By default, \Gls{sb3} converts the \Gls{gym} space \emph{Discrete} into one-hot vectors \citep{stable_baselines_2026}.

However, in some cases it is beneficial to have an ordinal relationship for discrete values. The implementation of the ordinal discrete space is achieved by using a continuous space (called \emph{Box}) and by mapping the discrete values to a continuous value between \(-1\) and \(1\).

\paragraph{Boolean and optional Boolean}
The Boolean space is implemented with a multi-binary space that consists of a single bit.
For the optional Boolean space, a categorical discrete space with \(3\) categories is used.

\begin{table}[hb]
    \centering
    \begin{tabular}{ll}
        \toprule
        Space Type & Values \\
        \midrule
        Categorical discrete & \(\{\vec{v} \in \{0, 1\}^n \mid \sum_{i=1}^{n} \vec{v}_i = 1\}\) \\
        Ordinal discrete & \(\{x \in \mathbb{R} | -1 \leq x \leq 1\}\) \\
        Multi discrete & \(\{ M \in \{0, 1\}^{m \times n} \mid \forall j \in \{1, 2, ..., n\}, \sum_{i=1}^{n} M_{ij} = 1\}\) \\
        Continuous & \(\mathbb{R}\) \\
        Boolean & \(\{0, 1\}\) \\
        Optional Boolean & \(\{\vec{v} \in \{0, 1\}^3 \mid \sum_{i=1}^{3} \vec{v}_i = 1\}\) \\
        \bottomrule
    \end{tabular}
    \caption{Observation Space Types}
    \label{tab:observation_space_types}
\end{table}

%% file: sections/appendix/reward_function.tex
\section{Reward Function}
\label{sec:reward_function}

During the reward shaping, several different reward functions were evaluated.
As a result, the version \(7\) performed best (Algorithm \ref{alg:reward_function_version_7}).
Note that the output is scaled by factor of \(\frac{1}{100}\) as a normalization measurement.
\begin{algorithm}[ht]
    \begin{algorithmic}[1]
        \State $\text{penalty}_{\text{depth}} \gets -\left(\frac{10}{\text{depth}_{\text{max}}}\right) \times \text{depth}$
        \If {$\text{invalid}$}
            \State \Return -10
        \EndIf
        \If {$\text{action} = \text{backtrack}$}
            \If {$\text{scored}_{\text{prev}} \neq \text{None} \ \textbf{and} \ \textbf{not} \ \text{scored}_{\text{prev}}$}
                \State \Return 15
            \Else
                \State \Return -10
            \EndIf
        \EndIf
        \If {$\text{operation} \ \textbf{is} \ \textbf{not} \  \text{scored}$}
            \State \Return 10 + $\text{penalty}_{\text{depth}}$
        \EndIf
        \If {$\text{score}$}
            \If {$\text{final}$}
                \State \Return 100
            \Else
                \State \Return 10 + $\text{penalty}_{\text{depth}}$
            \EndIf
        \EndIf
        \If {$\text{final}$}
            \State \Return -20 + $\text{penalty}_{\text{depth}}$
        \EndIf
        \State \Return -10 + $\text{penalty}_{\text{depth}}$
    \end{algorithmic}
    \caption{Reward Function Version \(7\)}
    \label{alg:reward_function_version_7}
\end{algorithm}

\FloatBarrier
\clearpage

%% file: sections/appendix/task_definitions.tex
\section{Task Definitions}
\label{sec:task_defintions}

\subsection{Task \task{sum list}}
The task \task{sum list} aims to calculate the sum of a list of single-digit integers.
The complexity is defined as the number of elements in the list.

\begin{figure}[h]
    \centering
     \includegraphics[width=\linewidth]{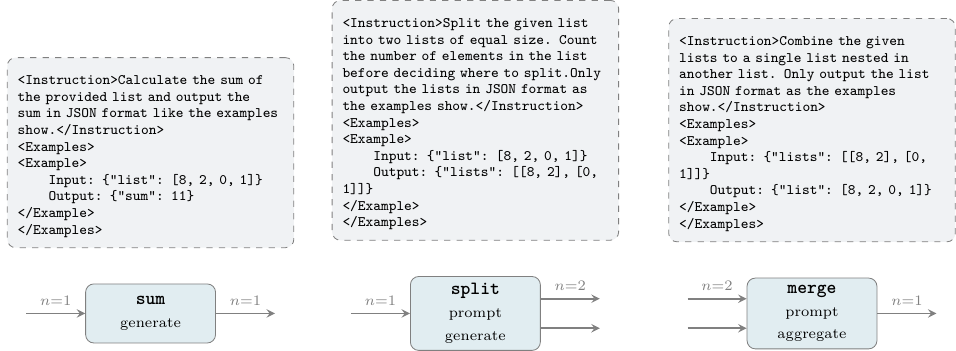}
    \caption{Operations of task \task{sum list} with input/output arities and prompt descriptions.}
\end{figure}

\paragraph{Operation \op{sum}}
For a given list of single-digit integers, the sum is to be calculated (Listing \ref{lst:task_sum_list_op_sum_definition}).
The scoring is based on whether the sum is correct.

\begin{figure}[hb]
\begin{lstlisting}[language=Python,tabsize=2,breaklines=true,frame=single,basicstyle=\tiny\ttfamily]
op_sum = PromptOperation(
    name='sum',
    n_outputs=1,
    n_inputs=1,
    type=OperationType.GENERATE,
    output_complexity=absolute_complexity(1),
    prompt=Prompt(
        instruction='Calculate the sum of the provided list and output the sum '
                    'in JSON format like the examples show.',
        examples=[
            Example(
                input={
                    'list': [8, 2, 0, 1]
                },
                output={
                    'sum': 11
                }
            )
        ]
    ),
    transform_before=lambda states: {
        'list': []
    } if len(states) == 0 else {
        'list': [states[0]['sum']]
    } if 'sum' in states[0] else states[0],
    score_operation=ScoreExecOperation(
        name='score_sum',
        type=OperationType.SCORE,
        score=score_op_sum,
        n_inputs=1,
        n_outputs=1
    )
)
\end{lstlisting}
    \caption{Definition of Operation \op{sum}}
    \label{lst:task_sum_list_op_sum_definition}
\end{figure}

\paragraph{Operation \op{split}}
A list is to be split into two equally sized sublists.
The scoring is based on whether the output contains two similarly sized lists, with some lenience (sublist ratio \(\geq 0.5\)).

\paragraph{Operation \op{merge}}
Two given lists are to be merged into a single list.
The result is scored based on whether the sum of the resulting list equals the sum of the input sublists and on the correct size of the merged list.

\FloatBarrier

\subsection{Task \task{sort list}}
The goal of the task \task{sort list} is to sort a list of single-digit integers.
The complexity is defined as the number of elements in the list.

\begin{figure}[h]
    \centering
     \includegraphics[width=\linewidth]{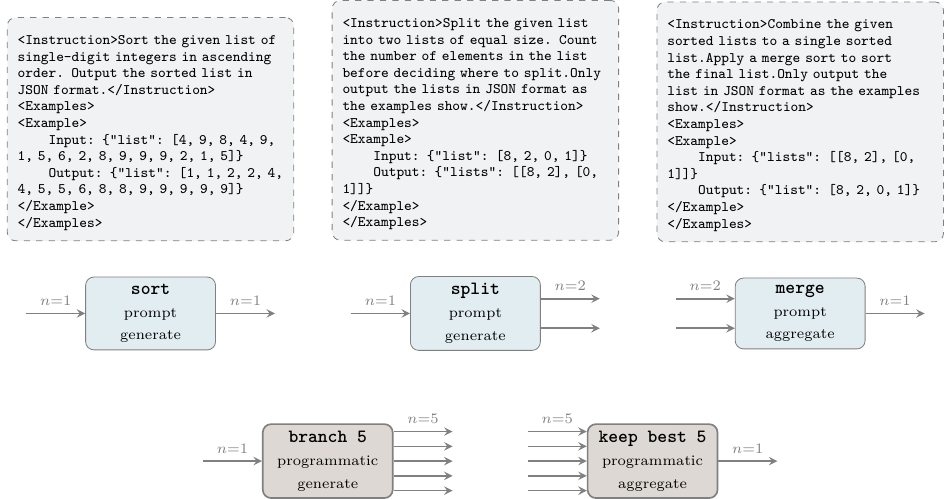}
    \caption{Operations of task \task{sort list} with input/output arities and prompt descriptions.}
\end{figure}

\paragraph{Operation \op{sort}}
A given list is to be sorted in ascending order.
The number of sort errors is calculated and the (binary) scoring strictly allows no errors.

\paragraph{Operation \op{split}}
Identical to the Operation \op{split} of Task \task{sum list}.

\paragraph{Operation \op{merge}}
Identical to the Operation \op{merge} of Task \task{sum list}.

\paragraph{Operation \op{branch 5}}
A programmatic operation that copies the state \(5\) times.

\paragraph{Operation \op{keep best from 5}}
A programmatic operation that selects the best of the incoming states.

\FloatBarrier

\clearpage

\subsection{Task \task{intersect set}}
This task aims to build the intersection of two sets.
The complexity is defined as the number of elements in the set.

\begin{figure}[h]
    \centering
     \includegraphics[width=0.75\linewidth]{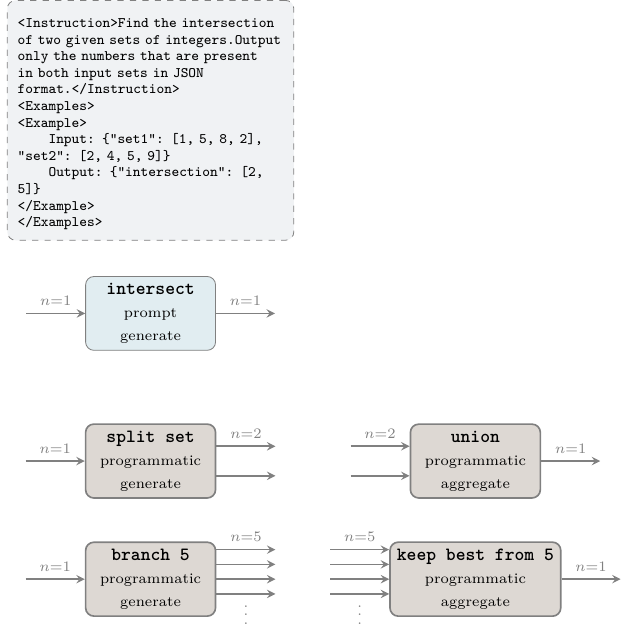}
    \caption{Operations of task \task{intersect set} with input/output arities and prompt descriptions.}
\end{figure}

\paragraph{Operation \op{intersect}}
The intersection of a set is to be created.
The scoring is based on whether the intersection of the sets is correct.

\paragraph{Operation \op{split set}}
Given two sets as input (a single state), two states are created:
The first set is split in half, the subsets are fed as first set into the respective output states, including the second (unmodified) set.

\paragraph{Operation \op{union}}
Given two states, the union of the sets is created.

\paragraph{Operation \op{branch 5}}
Copies the state \(5\) times.

\paragraph{Operation \op{keep best from 5}}
Selects the best state from given \(5\) states.

\FloatBarrier
\clearpage

\subsection{Task \task{count keywords}}
The goal is to count the keywords in a given text.
With the type of keywords defined (here: countries), the language model is advised to count the number of occurrences of the keywords (count per keyword).
The complexity is defined as the number of words in the text to count keywords of.

\begin{figure}[h]
    \centering
     \includegraphics[width=\linewidth]{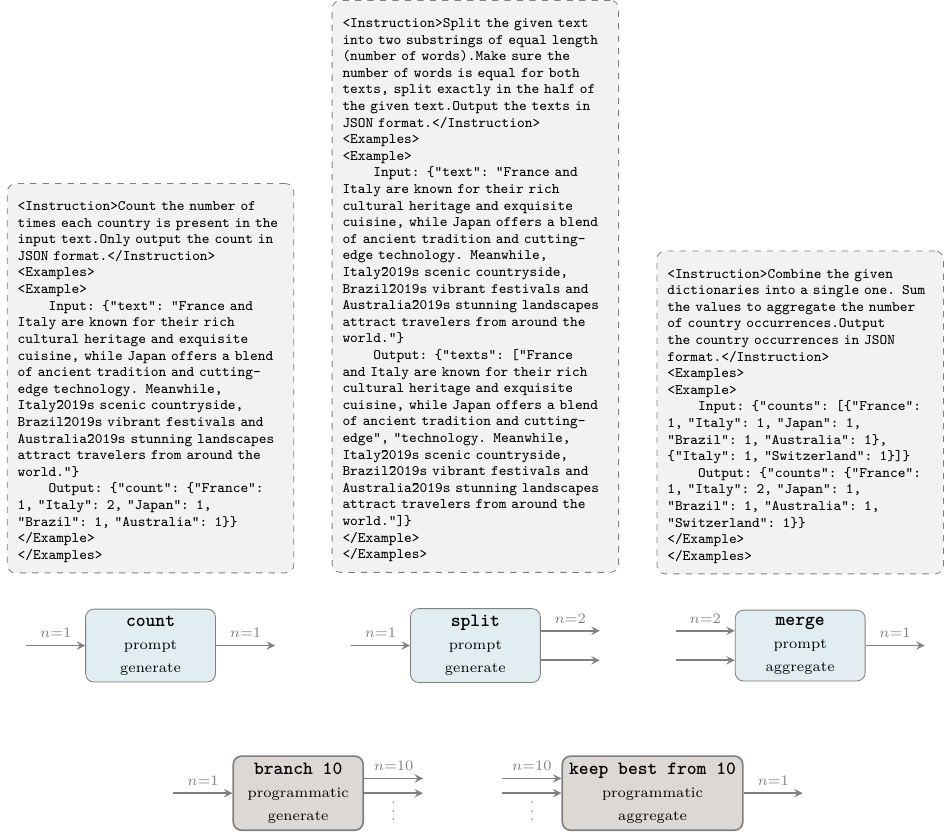}
    \caption{Operations of task \task{count keywords} with input/output arities and prompt descriptions.}
\end{figure}

\paragraph{Operation \op{count}}
Given a text, the number of specific words (here: countries) is to be counted.
The scoring is based on whether the words are counted correctly.

\paragraph{Operation \op{split}}
Given a text, it is to be split into two substrings of similar size.
Here, the scoring is dependent on whether the text is split correctly and in two similarly sized parts, given some lenience (text ratio \(\geq 0.25\)).

\paragraph{Operation \op{branch 10}}
Copies the state \(10\) times.

\paragraph{Operation \op{keep best from 10}}
Selects the best state from given \(10\) states.

\FloatBarrier

\subsection{Task \task{merge docs}}
The goal of \task{merge docs} is to merge a set of documents into a single document.
The complexity is defined as the number of documents to be merged.

\begin{figure}[h]
    \centering
     \includegraphics[width=0.75\linewidth]{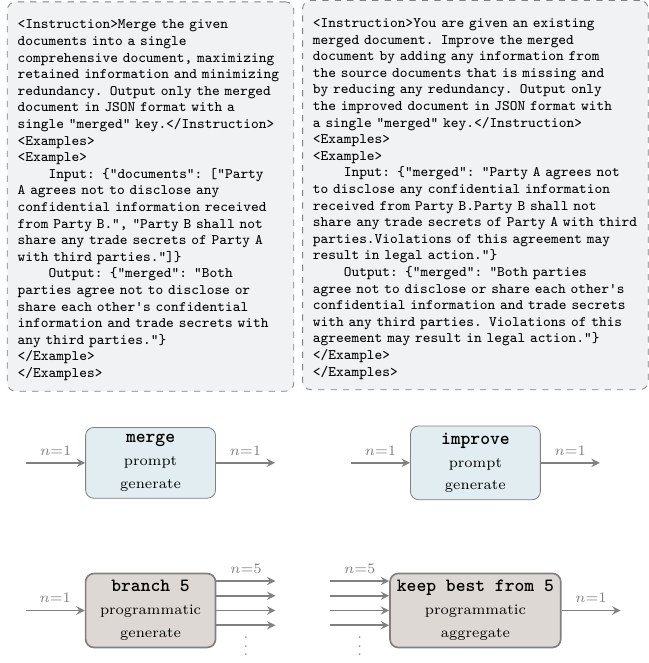}
    \caption{Operations of task \task{merge docs} with input/output arities and prompt descriptions.}
\end{figure}

\paragraph{Operation \op{merge}}
Several documents are to be merged into a single one, while minimizing redundancy and maximizing information retention.
The scoring is based on the ROUGE-L (retention) and ROUGE-1 (redundancy) score and the resulting F1 score \cite{rouge_lin_2004}.
To decide whether a result is valid, we used the threshold \(0.6\), which was determined empirically \footnote{Using the F1 threshold of \(0.6\) for the \op{merge} operation, language model evaluations showed a linear decrease of valid results with an increase of the number of documents to merge.}.
In practice, this threshold can be modified based on the use case.

\paragraph{Operation \op{improve}}
A merged document is to be improved in terms of information retention and redundancy minimization.
The scoring is identical to the \op{merge} operation.
Since we only have one document, the only complexity to be seen is \(1\).

\paragraph{Operation \op{branch 5}}
Copies the state \(5\) times.

\paragraph{Operation \op{keep best from 5}}
Selects the best state from given \(5\) states.

\FloatBarrier
\clearpage

%% file: sections/appendix/llm_capability_evaluation_results.tex
\section{\texorpdfstring{\acrshort{llm}}{LLM} Capability Evaluation}
\label{sec:llm_capability_evaluation_results}

\begin{figure}[hb]
    \centering
    \subfloat{%
        \resizebox{0.48\linewidth}{!}{\input{results/llm_capability_evaluation/sort_list/sort_list_all_ops}}%
    }
    \hfill
    \subfloat{%
        \resizebox{0.48\linewidth}{!}{\input{results/llm_capability_evaluation/intersect_set/intersect_set_all_ops}}%
    }
    \hfill
    \subfloat{%
        \resizebox{0.48\linewidth}{!}{\input{results/llm_capability_evaluation/merge_docs/merge_docs_all_ops}}%
    }

    \caption{Operation success probability \(\hat{P}(c)\) as a function of
             complexity \(c\) for each task, evaluated with
             \texttt{gpt-3.5-turbo-0125}.}
    \label{fig:llm_capability_evaluation_appendix}
\end{figure}
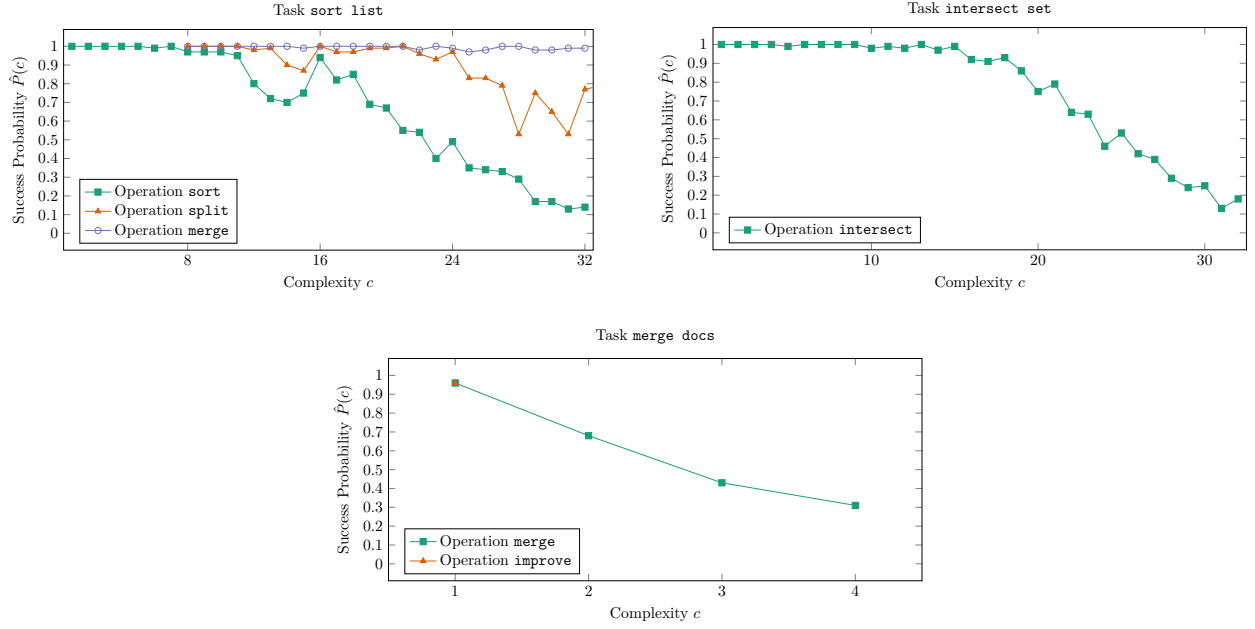

\subsection{Initial \texorpdfstring{\acrshort{llm}}{LLM} Evaluation}
\label{sec:additional_llm_evaluation}

\begin{figure}[hb]
    \centering
    \resizebox{0.7\textwidth}{!}{
        \input{./results/llm_capability_evaluation/sum_list/op_sum/additional_llm_capability_evaluation}
    }
    \caption{\acrshort{llm} Evaluation Results for Operation \op{sum} of Task \task{sum list} (with \(n=100\))}
    \label{fig:llm_capability_evaluation_additional_results}
\end{figure}
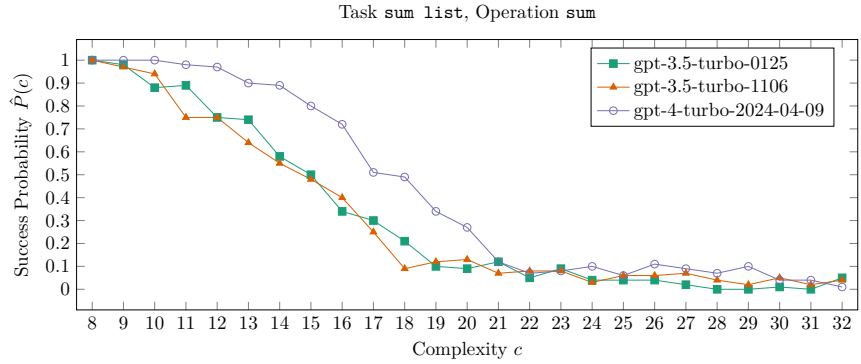

The operation \op{sum} of the task \task{sum list} was initially evaluated on different language models.
The variants of \llm{gpt-3.5-turbo} are quite on-par in solving the specified task. The model \llm{gpt-4-turbo-2024-04-09} maintains a higher success rate before degrading similarly (Figure \ref{fig:llm_capability_evaluation_additional_results}).

\clearpage

\section{Post-Hoc \texorpdfstring{\acrshort{llm}}{LLM} Evaluation}
\label{sec:post_hoc_llm_evaluation}

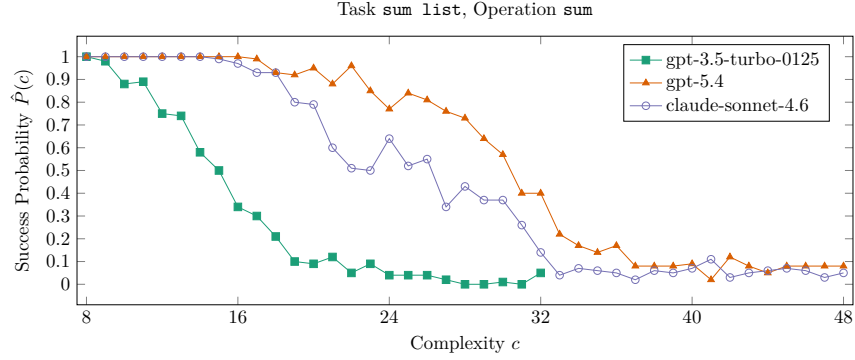
\begin{figure}[hb]
    \centering
    \resizebox{0.7\textwidth}{!}{
        \input{./results/llm_capability_evaluation/sum_list/op_sum/post_hoc_llm_capability_evaluation}
    }
    \caption{\acrshort{llm} Evaluation Results for Operation \op{sum} of Task \task{sum list} (with \(n=100\))}
    \label{fig:llm_capability_evaluation_post_hoc_results}
\end{figure}

To verify that recent \acrshortpl{llm} also show a degradation with an increase of a task's complexity, two modern models were evaluated after the training of the \acrshort{rl} agents (Figure \ref{fig:llm_capability_evaluation_post_hoc_results}).
We extended the evaluation up to list size \(48\) (\(50\%\) more than before), as the assumption was that the newer models perform better.
The results clearly show that while the models can keep up a higher success rate longer, they still degrade after a certain number of list elements.

%% file: results/llm_capability_evaluation/sort_list/sort_list_all_ops.tex
\begin{tikzpicture}
    \begin{axis}[
        title={Task \task{sort list}},
        xtick distance=8,
        ytick distance=0.1,
        ymin=0, ymax=1,
        xmax=32,
        enlarge x limits={abs=0.5},
        enlarge y limits={abs=0.09},
        scale only axis,
        xlabel={Complexity \(c\)},
        ylabel={Success Probability \(\hat{P}(c)\)},
        height=5cm,
        width=12cm,
        legend pos=south west,
        legend cell align={left},
        cycle list name=mark categories,
    ]
        \addplot table [x=cardinality, y=probability, col sep=comma]
            {results/llm_capability_evaluation/sort_list/op_sort/gpt35turbo0125.csv};
        \addlegendentry{Operation \op{sort}}
        
        \addplot table [x=cardinality, y=probability, col sep=comma]
            {results/llm_capability_evaluation/sum_list/op_split/gpt35turbo0125.csv};
        \addlegendentry{Operation \op{split}}

        \addplot table [x=cardinality, y=probability, col sep=comma]
            {results/llm_capability_evaluation/sum_list/op_merge/gpt35turbo0125.csv};
        \addlegendentry{Operation \op{merge}}
    \end{axis}
\end{tikzpicture}

%% file: results/llm_capability_evaluation/intersect_set/intersect_set_all_ops.tex
\begin{tikzpicture}
    \begin{axis}[
        title={Task \task{intersect set}},
        xtick distance=10,
        ytick distance=0.1,
        ymin=0, ymax=1,
        xmax=32,
        enlarge x limits={abs=0.5},
        enlarge y limits={abs=0.09},
        scale only axis,
        xlabel={Complexity \(c\)},
        ylabel={Success Probability \(\hat{P}(c)\)},
        height=5cm,
        width=12cm,
        legend pos=south west,
        legend cell align={left},
        cycle list name=mark categories,
    ]
        \addplot table [x=cardinality, y=probability, col sep=comma]
            {results/llm_capability_evaluation/intersect_set/op_intersect/gpt35turbo0125.csv};
        \addlegendentry{Operation \op{intersect}}
    \end{axis}
\end{tikzpicture}

%% file: results/llm_capability_evaluation/merge_docs/merge_docs_all_ops.tex
\begin{tikzpicture}
    \begin{axis}[
        title={Task \task{merge docs}},
        xtick distance=1,
        ytick distance=0.1,
        ymin=0, ymax=1,
        xmax=4,
        enlarge x limits={abs=0.5},
        enlarge y limits={abs=0.09},
        scale only axis,
        xlabel={Complexity \(c\)},
        ylabel={Success Probability \(\hat{P}(c)\)},
        height=5cm,
        width=12cm,
        legend pos=south west,
        legend cell align={left},
        cycle list name=mark categories,
    ]
        \addplot table [x=cardinality, y=probability, col sep=comma]
            {results/llm_capability_evaluation/merge_docs/op_merge/gpt35turbo0125.csv};
        \addlegendentry{Operation \op{merge}}
        
        \addplot table [x=cardinality, y=probability, col sep=comma]
            {results/llm_capability_evaluation/merge_docs/op_improve/gpt35turbo0125.csv};
        \addlegendentry{Operation \op{improve}}
    \end{axis}
\end{tikzpicture}

%% file: results/llm_capability_evaluation/sum_list/op_sum/additional_llm_capability_evaluation.tex
\begin{tikzpicture}
    \begin{axis}[
        title={Task \task{sum list}, Operation \op{sum}},
        xtick distance=1,
        ytick distance=0.1,
        ymax=1,
        xmax=32,
        enlarge x limits={abs=0.5},
        enlarge y limits={abs=0.09},
        scale only axis,
        xlabel={Complexity \(c\)},
        ylabel={Success Probability \(\hat{P}(c)\)},
        height=5cm,
        width=0.875\textwidth,
        legend pos=north east,
        legend cell align={left},
        cycle list name=mark categories
    ]
        \addplot table [
            x=cardinality, 
            y=probability,
            col sep=comma
        ] {results/llm_capability_evaluation/sum_list/op_sum/gpt35turbo0125.csv};
        \addlegendentry{gpt-3.5-turbo-0125}
        
        \addplot table [
            x=cardinality, 
            y=probability,
            col sep=comma
        ] {results/llm_capability_evaluation/sum_list/op_sum/gpt35turbo1106.csv};
        \addlegendentry{gpt-3.5-turbo-1106}
        
        \addplot table [
            x=cardinality, 
            y=probability,
            col sep=comma
        ] {results/llm_capability_evaluation/sum_list/op_sum/gpt4turbo20240409.csv};
        \addlegendentry{gpt-4-turbo-2024-04-09}
    
    \end{axis}
\end{tikzpicture}

%% file: results/llm_capability_evaluation/sum_list/op_sum/post_hoc_llm_capability_evaluation.tex
\begin{tikzpicture}
    \begin{axis}[
        title={Task \task{sum list}, Operation \op{sum}},
        xtick distance=8,
        ytick distance=0.1,
        ymax=1,
        xmin=8,
        xmax=48,
        enlarge x limits={abs=0.5},
        enlarge y limits={abs=0.09},
        scale only axis,
        xlabel={Complexity \(c\)},
        ylabel={Success Probability \(\hat{P}(c)\)},
        height=5cm,
        width=0.875\textwidth,
        legend pos=north east,
        legend cell align={left},
        cycle list name=mark categories
    ]
        \addplot table [
            x=cardinality, 
            y=probability,
            col sep=comma
        ] {results/llm_capability_evaluation/sum_list/op_sum/gpt35turbo0125.csv};
        \addlegendentry{gpt-3.5-turbo-0125}
        
        \addplot table [
            x=cardinality, 
            y=probability,
            col sep=comma
        ] {results/llm_capability_evaluation/sum_list/op_sum/gpt54.csv};
        \addlegendentry{gpt-5.4}
        
        \addplot table [
            x=cardinality, 
            y=probability,
            col sep=comma
        ] {results/llm_capability_evaluation/sum_list/op_sum/claudesonnet46.csv};
        \addlegendentry{claude-sonnet-4.6}
    \end{axis}
\end{tikzpicture}

%% file: sections/appendix/agent_evaluation_results.tex
\subsection{Agent Evaluation Results}
\label{sec:agent_evaluation_results}

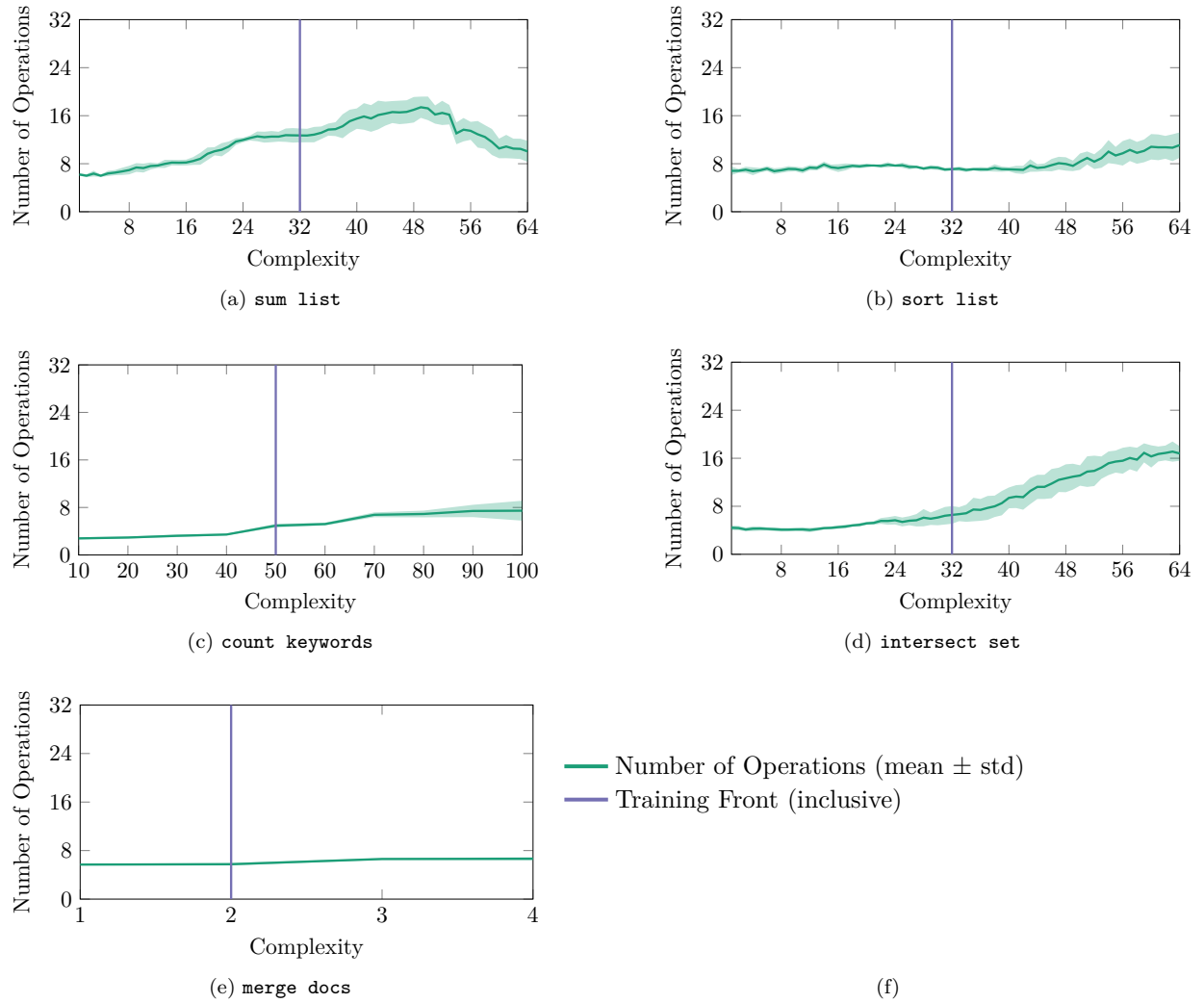
\begin{figure}[hb]
\centering
    \subfloat[\task{sum list}]{%
        \includegraphics[width=0.45\linewidth]{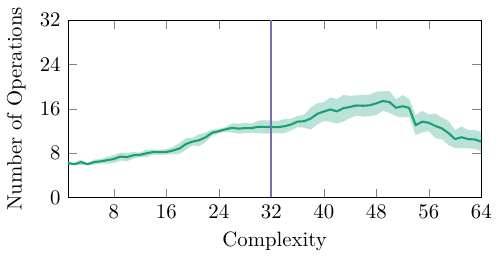}
    }\hfill
    \subfloat[\task{sort list}]{%
        \includegraphics[width=0.45\linewidth]{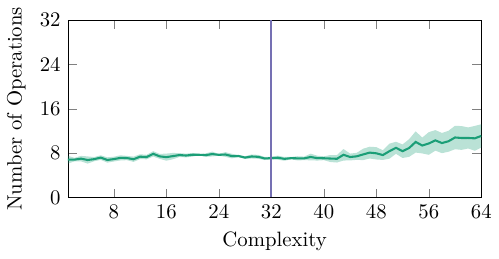}
    }\hfill
    \subfloat[\task{count keywords}]{%
        \includegraphics[width=0.45\linewidth]{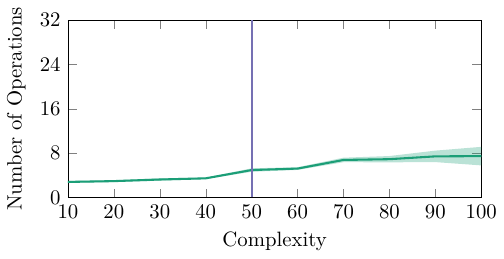}
    }\hfill
    \subfloat[\task{intersect set}]{%
        \includegraphics[width=0.45\linewidth]{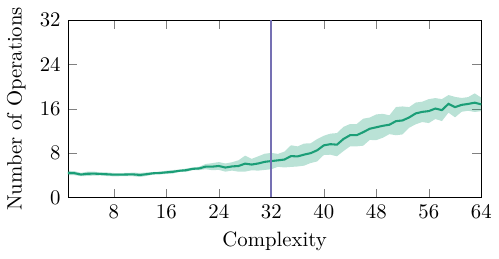}
    }\hfill
    \subfloat[\task{merge docs}]{%
        \includegraphics[width=0.45\linewidth]{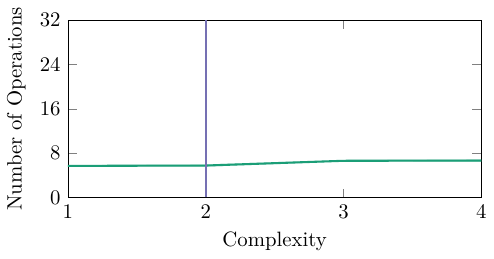}
    }
    \subfloat[]{%
        \begin{tikzpicture}
            \begin{axis}[
                hide axis,
                width=0.4\textwidth,
                height=0.4\textwidth,
                legend columns=1,
                legend style={draw=none, anchor=east, at={(-0.5,0.5)}},
                legend cell align={left},
                xmin=0, xmax=1, ymin=0, ymax=1,
            ]
                \addplot [draw=none, forget plot] coordinates {(0,0)};
        
                \addlegendimage{color=category a, line width=1.5pt}
                \addlegendentry{Number of Operations (mean $\pm$ std)}
                \addlegendimage{color=category c, line width=1.5pt}
                \addlegendentry{Training Front (inclusive)}
            \end{axis}
        \end{tikzpicture}%
    }\hfill
    \caption{Number of operations per complexity for all five tasks. The vertical line marks the training front, results to its right are out-of-distribution.}
    \label{fig:auto_got_n_operations}
\end{figure}

%% file: sections/appendix/ablation_summary.tex
\section{Ablation Summary}
\label{sec:ablation_summary}

In the expansion phase, we conducted a search over \acrshort{ppo} hyperparameters.
The findings are summarized below.

\paragraph{Clip range}
We introduced a clip range schedule annealing from \(0.15\) to \(0.3\) over the training.

A fixed low clip caused destructive early policy updates; the schedule reduces seed sensitivity significantly.
Values above \(0.3\) showed no benefit.

\paragraph{Update epochs}
We evaluated \(K \in \{5,8,10\}\) epochs per rollout batch. \(K = 8\) was optimal.
\(K = 10\) introduced instability on some tasks with negligible gain elsewhere.

\paragraph{Learning rate}
Among the schedules compared, linear decay performed better than cosine decay.
A warmup over the first \(10\%\) was added, dampening the impact of noisy early gradient steps.

Final schedule: linear decay from \(\alpha_0 = 5 \times 10^{-4}\) to \(\alpha_T = 1 \times 10^{-6}\) with warmup \(10\%\).

\paragraph{Total timesteps}
\(2^{18} = 262{,}144\) timesteps was used throughout.
Doubling to \(2^{19}\) caused collapse on several tasks, potentially due to the learning rate schedule decaying too slowly over the longer period.

\paragraph{Network architecture}
Compared \([64, 64]\) and \([128, 64]\) for both policy and value networks.
There was no significant benefit of employing the larger network, therefore \([64, 64]\) represents the final configuration.

\paragraph{Entropy coefficient}
Set to \(0.03\) to maintain sufficient exploration throughout training without destabilizing the policy.